# GeoConformal Prediction: a model-agnostic framework for measuring the uncertainty of spatial prediction

Xiayin Lou[a], Peng Luo[b]*, Liqiu Meng[a]

[a]*Chair of Cartography and Visual Analytics, Technical University of Munich, Munich, Germany;* [b]*Senseable City Lab, Massachusetts Institute of Technology, Cambridge, USA*

Spatial prediction is a fundamental task in geography, providing essential data support for various scenarios. Recent advancements, empowered by the development of geospatial artificial intelligence (GeoAI), have primarily focused on improving prediction accuracy while overlooking reliable measurements of prediction uncertainty. Such measures are crucial for enhancing model trustworthiness and supporting responsible decision-making. To address this issue, we propose a model-agnostic uncertainty assessment method called GeoConformal Prediction (GeoCP). First, a simulation study is conducted to validate the usefulness of GeoCP. Then, we applied GeoCP to two classic spatial prediction cases, spatial regression and spatial interpolation, to evaluate its reliability. For the case of spatial regression, we used XGBoost to predict housing prices, followed by GeoCP to calculate uncertainty. Our results show that GeoCP achieved a coverage rate of 93.67 %, while Bootstrapping methods reached a maximum coverage of 81.00 % after 2000 runs. We then applied GeoCP for the case of spatial interpolation models. By comparing a GeoAI-based geostatistical model with a traditional geostatistical model (Kriging), we found that the uncertainty obtained from GeoCP aligned closely with the variance in Kriging. Finally, using GeoCP, we analyzed the sources of uncertainty in spatial prediction. We found that explicitly including local features in AI models can significantly reduce prediction uncertainty, especially in areas with strong local dependence. Our findings suggest that GeoCP holds substantial potential not only for geographic knowledge discovery but also for guiding the design of future GeoAI models, paving the way for more reliable and interpretable spatial prediction frameworks.

## Introduction

Spatial prediction, as one of the core tasks in geography, has long attracted extensive attention from both academic and practical fields. Accurate prediction of spatial distribution of geographic variables provides essential support for natural resource management (Zuo and Xu 2023) , urban planning (Darabi et al. 2022), and environmental monitoring (Liu et al. 2020). Numerous models and algorithms have been developed to enhance the accuracy of spatial prediction (Luo 2024). Traditional methods primarily include Kriging (Luo et al. 2023) and spatial regression-based models (Sachdeva et al. 2023; Lessani and Li 2024), which have demonstrated unique value in addressing practical problems. In recent years, artificial intelligence (AI) technologies have been widely used for spatial prediction, with promising performance. For instance, deep neural networks can automatically learn complex patterns in spatial data, achieving high-precision predictions of geographic variables (Hagenauer and Helbich 2022; Chen et al. 2024). Large language models, leveraging their robust contextual understanding capabilities, can extract valuable information from unstructured data and incorporate it into spatial predictions (Guo et al. 2024; C. Liu et al. 2024).

In addition to the pursuit of prediction accuracy, another crucial aspect in the design of spatial prediction models is the robust measurement of prediction uncertainty (Luo et al. 2024). Only with an accurate understanding and assessment of the uncertainty associated with predictive models and their outcomes can spatial prediction models achieve greater interpretability and credibility. Moreover, a deeper understanding of uncertainty and the derivation of prediction results with confidence levels can improve model design and reduce bias in the process. This enhancement is essential to support reliable and responsible spatial decision-making in real-world applications (Zajko 2022). By quantifying and communicating uncertainty, scientists can develop spatial models that offer more credible insights, enabling decision-makers to assess risks and make more informed choices based on model confidence in its predictions. Here are several ways to define uncertainty. From the statistical perspective, uncertainty represents the degree of variability in predicted outputs due to model design, incomplete information, and randomness inside the datasets (McKay 1995). From a decision-theoretic viewpoint, uncertainty usually means the lack of certainty about outcomes, potential risks, or the best course of action because of the lack of information or knowledge about the situation (Scholz 1983). In social science, uncertainty reflects an individual's experience of doubt or unpredictability regarding future outcomes (Fiske 1991).

Some methods have been developed to measure the prediction uncertainty in the geospatial context. Hengl et al. (2017) use Shannon entropy to calculate the level of ambiguity of prediction results about the soil properties at each grid. Nevertheless, Shannon entropy depends on accurate probability estimation and is hard to interpret compared to prediction intervals and variance, etc. Spatial sensitivity analysis (Lilburne and Tarantola 2009; Xu and Zhang 2013; Saint-Geours et al. 2014) is another method for studying uncertainty. It quantifies how much the model output changes as the model inputs generate small variations while being hampered by the high computational costs. The prediction interval computed using cross-validation is also regarded as an indicator of uncertainty (Poggio et al. 2021) , but the prediction interval from cross-validation makes it difficult to guarantee 90% coverage, and does not reflect the actual uncertainty (Bates, Hastie, and Tibshirani 2024). Geostatistical methods (e.g., Ordinary Kriging) can quantify uncertainty by calculating the variance at each observation point. However, Kriging variance does not always accurately represent uncertainty (Heuvelink and Pebesma 2002). It is valid under certain circumstances, that is, the regionalized variable under study is a realization of a stationary Gaussian random function. Moreover, the variance calculation in Kriging is derived from the semivariogram, making it challenging to transfer this uncertainty quantification approach to other types of spatial prediction models. Consequently, there is no general uncertainty assessment framework applicable to various spatial prediction models currently.

In summary, although uncertainty quantification has received much attention, existing models still suffer from several drawbacks. Many approaches are computationally prohibitive (Lakshminarayanan, Pritzel, and Blundell 2017), may involve complex sampling strategies (Welling and Teh 2011; Janssen 2013), may require the modification to the model architecture (Gal and Ghahramani 2016), and may fail to handle covariate-shift problems (Chan et al. 2020; Flovik 2024). This makes it difficult to compare the uncertainty across different prediction tasks. Therefore, there is a pressing need for a model-agnostic, universal method that can be applied to any spatial prediction task to assess uncertainty.

Conformal prediction is a statistical learning method widely applied in uncertainty quantification within machine learning. The idea of conformal prediction is to use past experience to determine precise levels of confidence in new predictions (Shafer and Vovk 2008). Conformal prediction only relies on a weak assumption of exchangeability and does not require specific assumptions about the data distribution. Also, it can provide statistically rigorous uncertainty sets/intervals for the predictions (Angelopoulos and Bates 2021). Conformal prediction has been widely applied to different fields, including drug discovery (Eklund et al. 2015; Alvarsson et al. 2021), image classification (K. Liu et al. 2024), time-series forecasting (Zaffran et al. 2022), finance and credit scoring (Romano 2022), and natural language processing (NLP) (Giovannotti and Gammerman 2021), etc.

The original form of conformal prediction, commonly referred to as full conformal prediction (full CP), provides finite-sample valid prediction sets by evaluating a conformity score for every possible candidate output at a test point and comparing it to those of the training data. While this approach guarantees exact marginal coverage, it is computationally intensive because it requires retraining or re-evaluating the model for each candidate prediction (Gasparin and Ramdas 2025). sLSCP (Mao, Martin, and Reich 2024) is one of the few spatial conformal prediction models based on full CP framework. Although it offers a more efficient implementation when using Kriging as the underlying predictor, its scalability to other spatial models remains limited due to the inherent computational burden of full conformal prediction. Moreover, this method is restricted to continuous-valued regression tasks.

To address the inefficiency of full CP, split conformal prediction (split CP) was proposed. It partitions the dataset into a proper training set (used to fit the predictive model) and a calibration set (used to compute nonconformity scores, typically as residuals). This framework greatly improves computational efficiency while still ensuring valid marginal coverage under mild assumptions. A few studies have adapted the concept of split CP to specific spatial prediction models. For instance, Diana et al. (2023) introduced a split conformal approach for geographically weighted regression using tailored nonconformity measures. However, there remains no model-agnostic conformal prediction framework tailored for geospatial tasks that can flexibly handle both regression and classification problems.

The motivation of this study is to fill this gap by extending split conformal prediction to the geospatial domain, aiming to develop a model-independent uncertainty quantification framework for spatial prediction tasks. However, it is difficult to be directly applied split CP in a geospatial prediction context. First, spatial heterogeneity results in the covariate shift problem, which means that training and test sets belong to different distributions as location changes. Covariate shift can negatively affect the performance of geospatial models (Raitoharju 2022) and is very common in geospatial data. This makes getting reliable uncertainty estimates for conformal prediction challenging. Second, conformal prediction can only provide fixed prediction intervals, which makes it difficult to capture spatially varying uncertainty.

To address the covariate shift problem in conformal prediction, it is essential to measure the similarity between the training and test sets. This similarity can then be used as a weighting factor to mitigate the bias introduced by statistical differences between the two sets. While measuring similarity between datasets can be challenging, Tobler's First Law of Geography provides a valuable framework in geographic contexts, helping to describe this issue effectively. Motivated by that, we propose GeoCP, introducing geographic weights as the similarity measure to alleviate the covariate shift problem. The introduction of geographic weights also makes it possible to estimate spatially varying uncertainty at different locations and compare the uncertainty among different datasets and models.

We validate the reliability of the uncertainty estimates obtained from GeoCP. First, we design a simulation study that artificially forms a true prediction model and biased prediction models with simulated uncertainty. The uncertainty predicted by GeoCP will be evaluated by comparing it with the well-designed uncertainty. Then, using real-world datasets, GeoCP is applied to two classical spatial prediction frameworks: spatial regression and spatial interpolation. Spatial regression is usually based on the correlation of the observed variable with other explanatory variables. Spatial interpolation is based on the autocorrelation of the observed variable itself, and the explanatory variables are usually absent.

In the spatial regression case, we combine a machine learning regression model with GeoConformal to predict the housing price and output the uncertainty. We first computed the prediction uncertainty in two situations: with and without spatial features. Then, we compared the prediction uncertainty of these two situations to investigate the impact of spatial features on the spatial prediction results. Next, we generated the error distribution of the spatial regression model using the bootstrapping method and compared it with the prediction uncertainty at different quantile levels.

In the spatial interpolation case, we selected two models: a geostatistical model, Ordinary Kriging (OK), and a GeoAI model, specifically the Deep Geometric Spatial Interpolation (DGSI) framework. DGSI follows the principles of Kriging but, instead of using a semivariogram, it predicts the weights of surrounding observations using a neural network framework. These two spatial interpolation models were chosen because they rely solely on the spatial distribution characteristics of the predicted variable without incorporating any explanatory variables. This ensures that the prediction results are entirely dependent on the spatial distribution of the target variable and the model itself, allowing a more straightforward evaluation of the reasonableness of the uncertainty estimates. We first explored the values and spatial characteristics of uncertainty across different models. Then, we analyzed the relationship between uncertainty and the spatial patterns of geographic variable distribution.

Finally, leveraging GeoCP, we investigated how different representations of spatial information affect the uncertainty in GeoAI prediction models. Building on DGSI, we developed two new models aimed at more explicitly capturing the local features of variables. We compared the uncertainty of these new models with that of the initial model, analyzing how the changes in uncertainty relate to the spatial distribution of the geographic variables themselves.

The remainder of the manuscript is organized as follows: In Section 2, we introduced the GeoConformal prediction method. In Section 3, we conducted a simulation study. Section 4 describes the experiment design for two spatial prediction tasks. Section 5 presents the spatial regression experiments, followed by the spatial interpolation experiments in Section 6. We explored the prediction uncertainty of different spatial explicit strategy in Section 7. We discussed the experimental results in Section 8, and conclude the study in Section 9.

## GeoConformal Prediction (GeoCP)

### *Conformal prediction and its challenges in spatial prediction*

Conformal prediction (Lei et al. 2018) is a simple but powerful framework for quantifying the uncertainty of any machine learning model. It converts a model's prediction output into prediction sets or intervals as a proxy of uncertainty (Vovk, Gammerman, and Shafer 2005). Given a model f: $\mathcal{X} \rightarrow \mathcal{Y}$ fitted on a training set and *m* additional data points $(X_1, y_1), \ldots (X_m, y_m)$ as a calibration set. With $f$ and calibration set, conformal prediction seeks to construct a prediction interval or set that is valid in the sense that

$$\mathbb{P}[y_{test} \in \mathcal{C}(X_{test})] \geq 1 - \varepsilon \tag{1}$$

where $(X_{test}, y_{test})$ is a new test point, $\varepsilon \in [0,1]$ is the user-defined miscoverage rate. In a nutshell, the probability of this prediction interval or set containing the correct value is almost exactly 1-$\varepsilon$. Then, the prediction set or interval for this new test data point is computed as follows:

$$\mathcal{C}(X_{test}) = \left\{ y: \alpha(f(X_{test}), y_{test}) \leq \text{Quantile}_{1-\varepsilon}\left(\frac{1}{m}\sum_{i=1}^{m} \delta_{\alpha_i}\right) \right\} \tag{2}$$

in which $\mathcal{C}(\cdot)$ is the prediction set or interval given by conformal prediction, $\delta_{\alpha_i}$ denotes a point mass at a value $\alpha_i$ (i.e., the distribution that places probability 1 on the value $\alpha_i$), $\frac{1}{m}\sum_{i=1}^{m}\delta_{\alpha_i}$ is a discrete probability distribution reflecting the empirical distribution of $y_1, \ldots, y_m$. $\alpha(\cdot)$ is the nonconformity score function, a measure of how well a new data point conforms to a model trained on a given dataset. For a regression task, $\alpha(\cdot)$ can be the absolute difference between the predicted value and the ground truth, which is the most straightforward nonconformity measure (Kato, Tax, and Loog 2023). For a classification task, $\alpha(\cdot)$ can be the inverse probability, which is also known as hinge loss function (Johansson et al. 2017). $\alpha_i$ is the nonconformity score for $i$th point in the calibration set. $\text{Quantile}_{1-\varepsilon}$ is the (1-ε)-quantile value of nonconformity scores from previously seen n calibration data points, ε is the miscoverage level.

Measuring uncertainty with spatial variability is crucial for understanding the performance of spatial predictions and offering insights for spatial decision-making. Conformal prediction assumes that training, calibration, and test sets are drawn from the same distribution or, more generally, $(X_1, y_1), \ldots (X_m, y_m), (X_{test}, y_{test})$ are exchangeable. This exchangeability assumption is slightly weaker than the independent and identically distributed (i.i.d.) assumption. However, the existence of spatial heterogeneity carries statistical challenges for conformal prediction, which is defined as the covariate shift problem (Tibshirani et al. 2019) , breaking the exchangeability assumption. Covariate shift means that the input (e.g., training and test set) distribution can change while the fitted model remains the same, thus degrading the performance of models. Due to the spatial heterogeneity of geographical variables, the spatial distribution of observations is very uneven across the space. So, the prediction uncertainty should vary across space. In addition, the spatially unbalanced sampling can also contribute to this problem. However, the original conformal prediction only provides an average coverage guarantee for the entire dataset, limiting its application in spatial prediction tasks (Han et al. 2022).

***Spatial dependence as the solution for covariate shift in Conformal prediction***

As mentioned earlier, a constant prediction interval computed by the original conformal prediction fails to offer an optimal uncertainty for each location. Given that the data points $(X_1, y_1), \dots (X_m, y_m), (X_{test}, y_{test})$ are no longer exchangeable due to covariate shift, we can relax the exchangeability assumption: the calibration data points $(X_1, y_1), \dots (X_m, y_m)$ are drawn from a distribution $P$ and test data point $(X_{test}, y_{test})$ from another distribution $Q$, but with the restriction that $P_{Y|X} = Q_{Y|X}$, that is, the conditional distribution of Y|X remains the same for both training and test sets. Under this setting, if the covariate likelihood ratio (or similarity) from the test to the training set is known, conformal prediction can still work by using the quantile of a suitably weighted empirical distribution of nonconformity scores (Tibshirani et al. 2019; Bhattacharyya and Barber 2024). This modified version of conformal prediction is called weighted conformal prediction (WCP), in which higher weights are assigned to data points that are 'trusted' more, namely, the test points and calibration points with similar features may come from a similar distribution (Barber et al. 2023).

The core of the weighted conformal prediction is to estimate the likelihood ratio of covariates between test and training datasets. As mentioned above, the covariate distributions of training and test sets are assumed to be different, so each nonconformity score $\alpha_i$ will be weighted by a probability proportional to the covariate likelihood ratio $\mathrm{w}(X_i)$. When the weight of each nonconformity score is $1/m$, the weighted conformal prediction will degrade into the original conformal prediction. The weighted conformal prediction is defined as follows.

$$\mathcal{C}_{\mathrm{w}}(X_{test}) = \left\{ y : \alpha(f(X_{test}), y_{test}) \leq \text{Quantile}_{1-\varepsilon}\left( \sum_{i=1}^{m} w_i(X_i) \cdot \delta_{\alpha_i} \right) \right\} \quad (3)$$

Where $w_i(X_i)$ captures the shift from the training set to the test set, $\delta_{\alpha_i}$ is the point mass at $\alpha_i$ . However, in the geospatial context, there are few variables (e.g., spatial interpolation). And the uncertainty computed may be nontransferable because the variables can change over datasets. In some spatial prediction contexts, there is a lack of sufficient explanatory variables. For example, this issue arises in spatial interpolation for mineral content, where predictions can only rely on sample data with coordinates. $\{u_i, v_i\}_{i=1}^n$ and values $\{z_i\}_{i=1}^n$. Hence, it is difficult for weighted conformal prediction to provide spatially varying uncertainty. In addition, if the weights are computed based on the similarity of features, then the estimated uncertainty may be influenced largely by the different selection of features and different datasets. Here, we can build a universal framework for estimating geographic uncertainty for various models and datasets by generating weights with only geographic locations rather than features. In fact, it may also unveil some patterns of geographic data.

Measures of distributional similarity across locations has been well discussed, and probably the most famous of these is Tobler's first law of geography, i.e., observations nearby have similar features. In other words, as the distance between the test data point and the calibration data points increases, the geographic weight decays. As a result, geographic weights are introduced to extend weighted conformal prediction, as shown in Figure 1, which can provide optimal prediction at each geographic location. A mathematical proof demonstrating that the distance-weighted split conformal prediction satisfies the local weighted exchangeability condition is included in Appendix C.

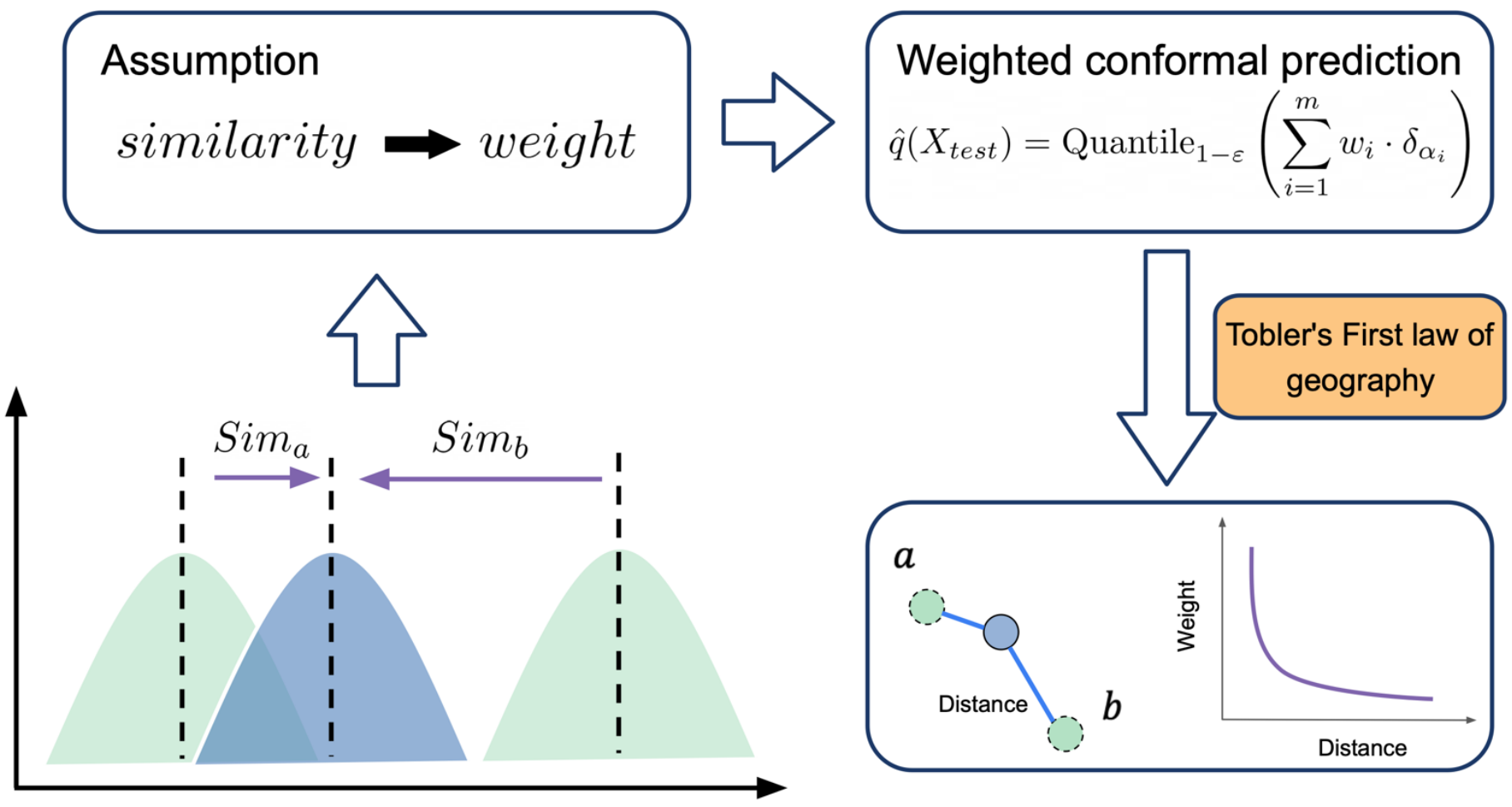

Figure 1 The first law of Geography as the guide to solve the covariate shift problem

Therefore, geographic weighting as a proxy of the likelihood ratio is introduced to the original conformal prediction. The prediction region for $X_{test}$ at geographic location $(u_{test}, v_{test})$ can be represented as follows:

$$\mathbb{P}\left[y_{test} \in C_{geo}(X_{test})\right] | (u_{test}, v_{test})] \geq 1 - \varepsilon \tag{4}$$

$$C_{geo}(X_{test}) = \{y: \alpha(f(X_{test}), y_{test}) \leq \text{GeoQuantile}_{1-\varepsilon}(u_{test}, v_{test}; \alpha_1, \dots, \alpha_m)\} \tag{5}$$

$$\text{GeoQuantile}_{1-\varepsilon}(u_{test}, v_{test}; \alpha_1, \dots, \alpha_m) = \text{Quantile}_{1-\varepsilon}\left(\sum_{i=1}^{m} w_i(u_{test}, v_{test}) \cdot \delta_{\alpha_i}\right) \tag{6}$$

where $\text{GeoQuantile}_{1-\varepsilon}(\cdot)$ is the geographically weighted (1- ε)-quantile function, $(u_{test}, v_{test})$ is the geographic location of $X_{test}$, $w_i(u_{test}, v_{test})$ is the geographic weight assigned to $i$th calibration data point. The pseudocode of calculating geographically weighted quantile for a test point is demonstrated in Algorithm 1.

**Algorithm 1**: Geographically Weighted Quantile

**Input:** Nonconformity scores $\{\alpha_i\}_{i\in\mathcal{I}_{calib}}$, calibration set $\{X_i, y_i\}_{i\in\mathcal{I}_{calib}}$, test point $X_{test}$, miscoverage level ε, geographic distance function dist(·), decay function β

**Output**: Geographically weighted quantile $\hat{q}$ for test point $X_{test}$.

m $\leftarrow \text{len}(\mathcal{J}_{calib})$

$W^{geo} \leftarrow$ List(m) // An empty list with length $m$

**for** i $= 1$ **to** m **do**

d $\leftarrow dist(X_{test}, X_i^{calib})$

$W_i^{geo} \leftarrow \beta(d)$

$\{\alpha_{(i)}, W_{(i)}^{geo}\}_{i=1}^{m} \leftarrow \text{sort}(\{\alpha_i, W_i^{geo}\})$ such that $\alpha_{(1)} \leq \alpha_{(2)} \leq \cdots \leq \alpha_{(m)}$

**for** i $= 1$ **to** m **do**

$W_{(i)}^{geo} \leftarrow W_{(i)}^{geo} / \sum_{j=1}^{m} W_{(j)}^{geo}$

$F_i = \sum_{j=1}^{i} W_{(j)}^{geo}$

k $\leftarrow$ smallest index such that $F_k \geq 1 - \varepsilon$

$\hat{q} \leftarrow \alpha_{(k)}$

***Framework of GeoConformal Prediction (GeoCP)***

In summary, to deal with the covariate shift problem and offer spatially varying uncertainty, we introduce geographic weights into conformal prediction. The procedure for GeoCP (see Figure 2) is outlined as four stages. To begin with, the dataset is split into a training set, a calibration set, and a test set. The calibration set is assumed to have same distribution as the training set. Next, a geospatial model f is fitted on the training set. Then, a series of nonconformity scores are computed based on the calibration set. The fourth stage can be summarized in the following steps. The nonconformity scores will first be ranked ascendingly. The geographic weights between each test data point and all calibration data points are calculated according to their geographic distances. The geographic weights decay as the distances increase. There are many different distance decay methods that can be used in GeoCP. In the simulation and real-world experiments presented in this paper, we adopted the Gaussian distance decay function, as previous studies have shown that it consistently yields lower average interval scores compared to other kernel functions (Mao, Martin, and Reich 2024). Finally, we determine the (1- ε)-level quantile position by finding the data value where the cumulative geographic weight equals or just surpasses the target quantile level. This (1- ε)-level geographically weighted quantile is the uncertainty for the test data points. The pseudocode of GeoCP is shown in the Algorithm 2.

**Algorithm 2**: GeoConformal Prediction

**Input:** Dataset $\{X_i, y_i\}_{i \in \mathcal{I}}$, geospatial model f: $\mathcal{X} \rightarrow \mathcal{Y}$, nonconformity score function α, miscoverage level ε.

**Output**: Prediction interval $\mathcal{C}_{geo}(X_{\mathrm{i}}^{test}), \mathrm{i} \in \mathcal{I}_{\mathrm{test}}$ for test points

**Stage 1:** Split dataset into training set $\{X_i^{train}, y_i^{train}\}_{i \in \mathcal{I}_{train}}$, calibration set $\{X_i^{calib}, y_i^{calib}\}_{i \in \mathcal{I}_{calib}}$, test set $\{X_i^{test}, y_i^{test}\}_{i \in \mathcal{I}_{test}}$.

**Stage 2:** Fit geospatial model $f$ on training set $\{X_i^{train}, y_i^{train}\}_{i \in \mathcal{I}_{train}}$.

**Stage 3:** Compute nonconformity scores using calibration set $\{\alpha_i\}_{i \in \mathcal{I}_{calib}}$, where $\alpha_i = \alpha\left(f\left(X_i^{calib}\right), y_i^{calib}\right)$.

Stage 4: Calculate geographically weighted quantile

$q_{\mathrm{i}}=\mathrm{GeoQuantile}_{1-\varepsilon}\left(u_i^{test}, v_i^{test}; \{\alpha_i\}_{i \in \mathcal{I}_{calib}}\right)$ and prediction intervals $\mathcal{C}_{geo}(X_{\mathrm{i}}^{test}) = \{y: \alpha(f(X_i^{test}), y_i^{test}) \leq q_i\}$ for each test point.

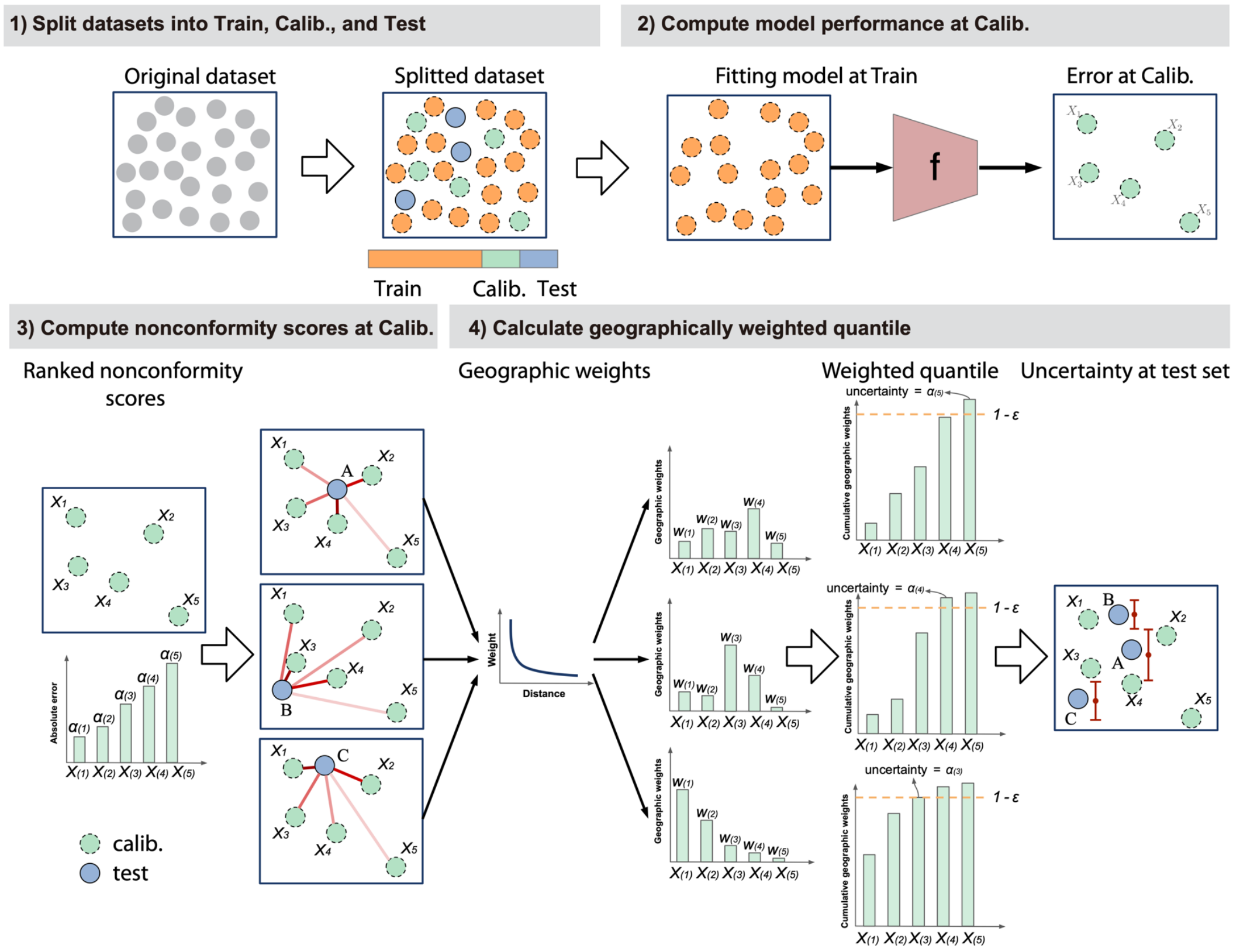

Figure 2 A framework of Geoconformal Prediction

## Simulation study

In this section, we designed the workflow to simulate the datasets with different spatial autocorrelation levels and validated the performance of the GeoCP for quantifying its uncertainty at different levels of autocorrelation.

### *The framework for evaluating uncertainty measurements with a biased model*

We developed a novel framework to evaluate the predicted uncertainty provided by GeoCP and other models for measuring prediction uncertainty, as shown in Figure 3.

First, we performed a Data Generation Process (DGP) based on a specific formula, obtaining the corresponding Y value at each specific location (u, v). Next, we defined the true model and biased model. The true model is a prediction model that strictly aligns with the data generation process. It accurately predicts the true Y value based on the input (u, v). The biased model, on the other hand, is built upon the true model. For each input (u, v), it generates an output with a simulated uncertainty $\Delta(u, v)$, represented as (true Y + $\Delta(u, v)$). In this way, we constructed a prediction model with a clearly defined uncertainty using the biased model.

Third, we input the generated data into the biased model and then used GeoCP to calculate the prediction uncertainty $\Delta(u, v)$ of the biased model. Finally, we followed the same process by inputting the data into the true model and using GeoCP to calculate the uncertainty. We compare the true uncertainty $\Delta(u, v)$ and predicted uncertainty $\widehat{\Delta}(u, v)$ using different metrics. Theoretically, the predicted uncertainty for the true model should be 0 because there is no uncertainty in true model.

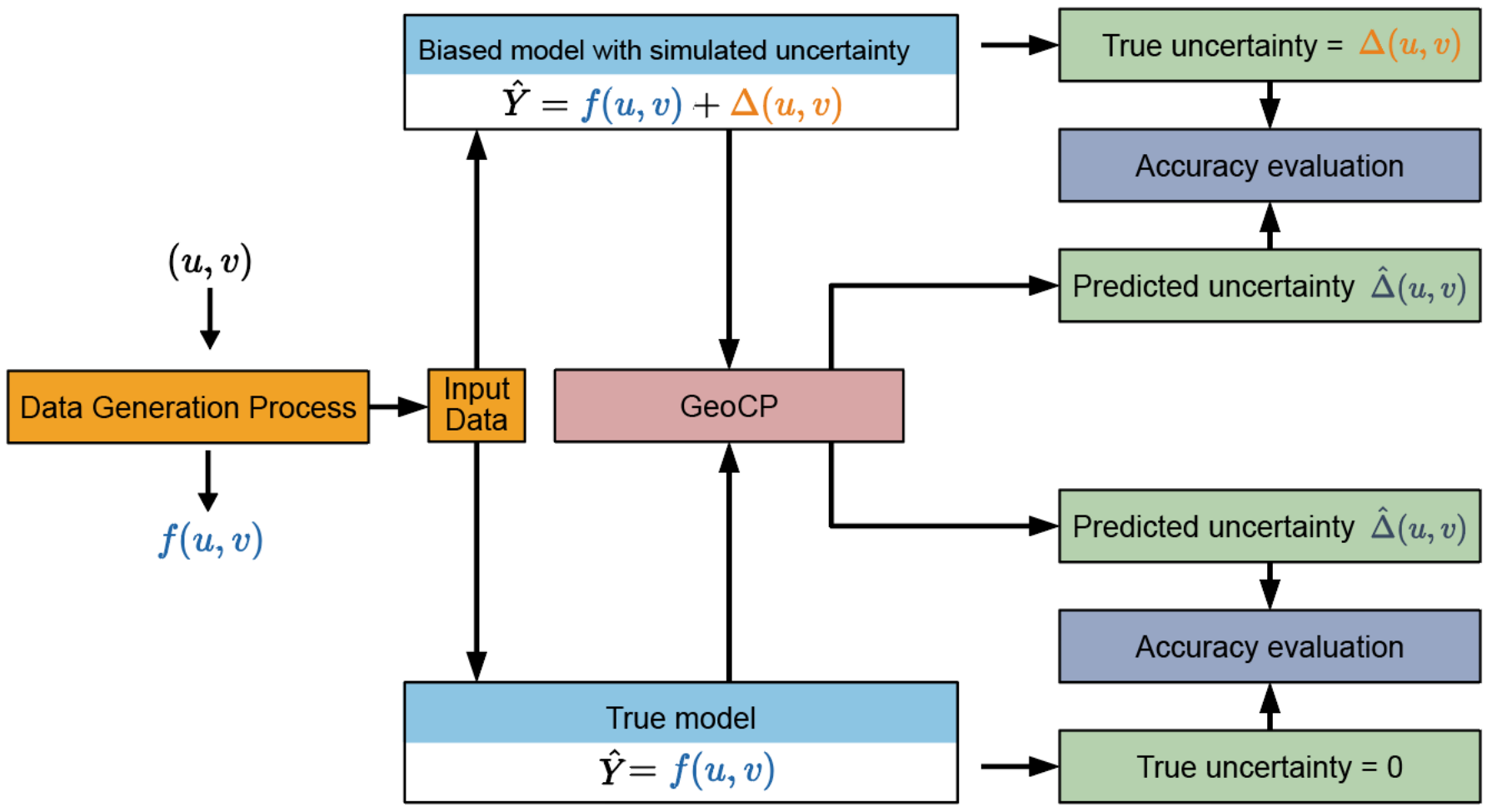

Figure 3 The workflow of accuracy evaluation of simulated uncertainty

***Experiment design for assessing the performance of GeoCP***

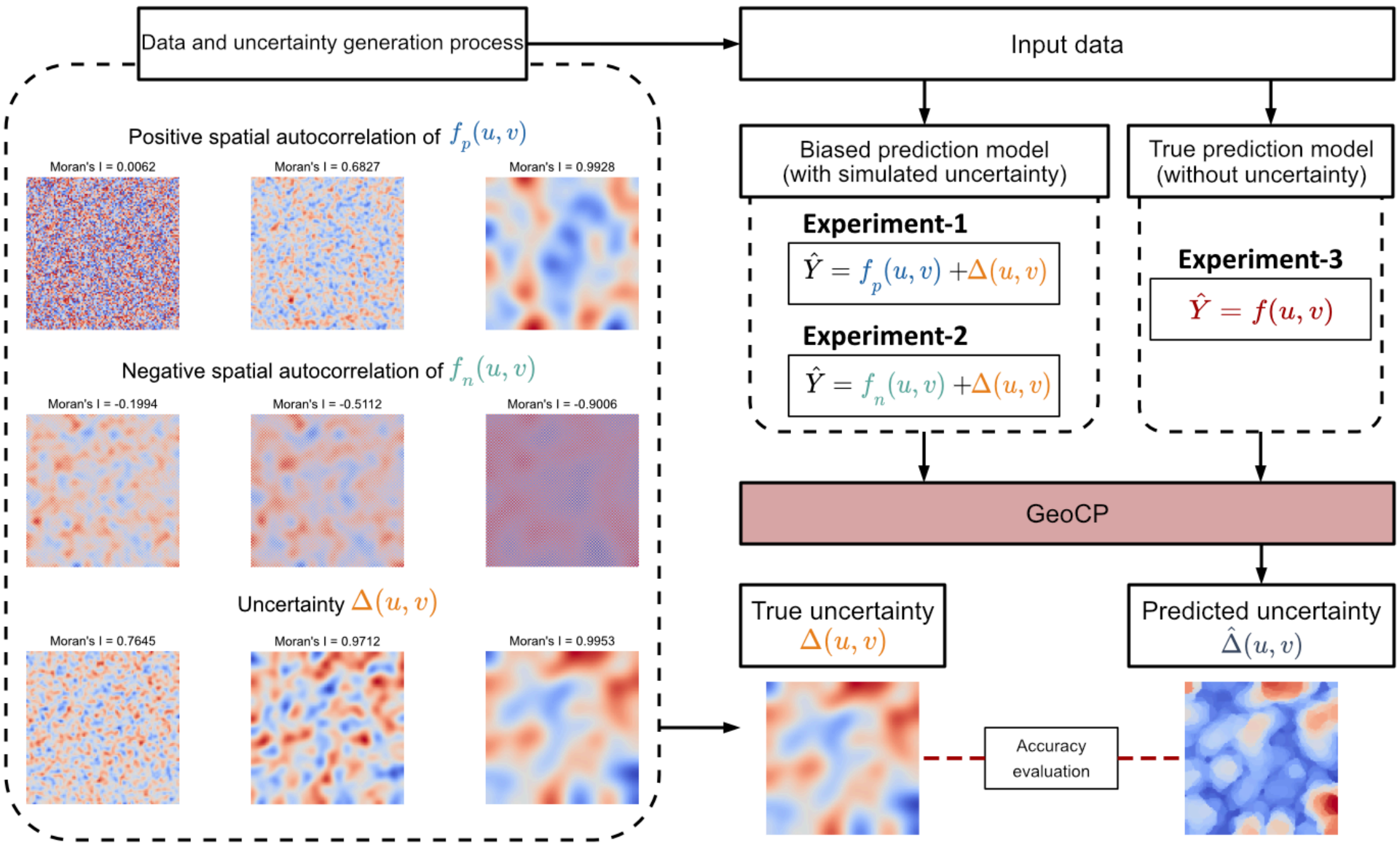

Figure 4 An overview of simulation design

Based on the uncertainty evaluation method described in the previous section, we conducted a series of experiments to assess the performance of GeoCP under different DGPs. Specifically, we aim to investigate how GeoCP is affected by varying spatial autocorrelation in both the data and uncertainty.

Through well-defined data generation processes, we can formulate models with known and controlled behaviours. With predefined ground truth, this method allows us to verify the ability of GeoCP to capture the uncertainty of models. Figure 4 shows the DGP and process of model validation. First, we simulated $Y$ with positive and negative spatial autocorrelation, which is defined as follows:

$$Y = f_p(u,v) = G\big(\beta_p(u,v)\,,\lambda\big) \text{ where } \beta_p(u,v) \sim U(10{,}20) \tag{7}$$

$$\begin{gathered} Y = f_n(u,v) = \beta_n(u,v) + G(\phi(u,v),\lambda) \text{ where} \\ \beta_n(u,v) = 15 + \big((u+v) \bmod 2\big), \phi(u,v) \sim U(-5{,}5) \end{gathered} \tag{8}$$

in which $(u, v)$ is one coordinate on a 100×100 grid, $G(Z, \lambda)$ is a 2d gaussian filter with $\lambda$ adjusting the smoothness of values, $U(a, b)$ is a uniform distribution with lower and upper boundaries of $a$ and $b$. Functions $\beta_p(u, v)$ and $\beta_n(u, v)$ are regarded as the base for building true models; $f_p(u, v)$ and $f_n(u, v)$ denote the functions that generate y with positive and negative spatial autocorrelation, respectively, $\phi(u, v)$ is used to alleviate the negative spatial autocorrelation. Output Y from these two functions $f_p(u, v)$ and $f_n(u, v)$ all range from 10 to 20. By changing the value of $\lambda$, we are able to adjust the level of spatial autocorrelation of these models. In this paper, we are interested in capturing spatial uncertainty in the models, so we formulated uncertainty $\Delta(u, v)$ with different levels of spatial patterns, which is defined as:

$$\Delta(u, v) = G(\beta_\Delta(u, v), \lambda) \text{ where } \beta_\Delta(u, v) \sim N(\mu, \sigma^2) \tag{9}$$

in which $\mu$ and $\sigma^2$ specify mean and variance of the normal distribution, $\beta_\Delta(u, v)$ is the base for generating uncertainty, which is drawn from a normal distribution $N(\mu, \sigma^2)$. As a result, a general form of biased $\hat{Y}$ can be framed as follows:

$$\hat{Y} = f(u, v) + \Delta(u, v) \tag{10}$$

where $f(u, v)$ can be either $f_p(u, v)$ or $f_n(u, v)$.

In this study, we designed three simulation experiments (right part of Figure 4), and the predicted uncertainty will be evaluated using different metrics.

***Results***

Figure 5 summarizes the results of simulation study. Here, two metrics are used: *normalized root mean square error* (NRMSE) and *coverage rate* (COVER). NRMSE measures the accuracy via comparing the normalized predicted values to the normalized true values. We employed this metric because changing the spatial autocorrelation will carry the magnitude change of uncertainty. COVER computes the percentage of true values that lie within the prediction interval.

It is obvious that the spatial pattern of the data has a trivial impact on the uncertainty predicted (see Figure 5a, b). The values of NRMSE and COVER don't change much along the vertical axis. As shown in Figure 6, with the same true uncertainty, the predicted uncertainty and accuracy metrics are almost the same even though the true models have totally opposed spatial autocorrelation. In contrast, the predicted uncertainty is influenced largely by spatial pattern of simulated uncertainty. Figure 5c, d demonstrates the relationship between the uncertainty spatial autocorrelation levels and the two metrics. As the spatial patterns (measured in Moran's I) of uncertainty become stronger, RMSE decreases, and COVER increase quadratically. This happens because GeoCP assumes that uncertainty is spatially varying. Take a closer look at the results, as illustrated in Figure 6a, b, the higher spatial autocorrelation the true uncertainty has, the better GeoCP can recover it. For the true prediction model (see Figure 6c), where no uncertainty exists in the prediction process, the predicted uncertainty from GeoCP is exactly 0. While Moran's I of true uncertainty (first row) is relatively high, the clusters are dispersed rather than forming compact regions. This makes it difficult for GeoCP to recover the true uncertainty. However, our proposed GeoCP method can still guarantee a coverage rate of over 86% even the spatial autocorrelation is close to 0, which means the uncertainty is randomly distributed. Moreover, in the three experiments, the magnitude of predicted uncertainty is roughly consistent with that of true uncertainty.

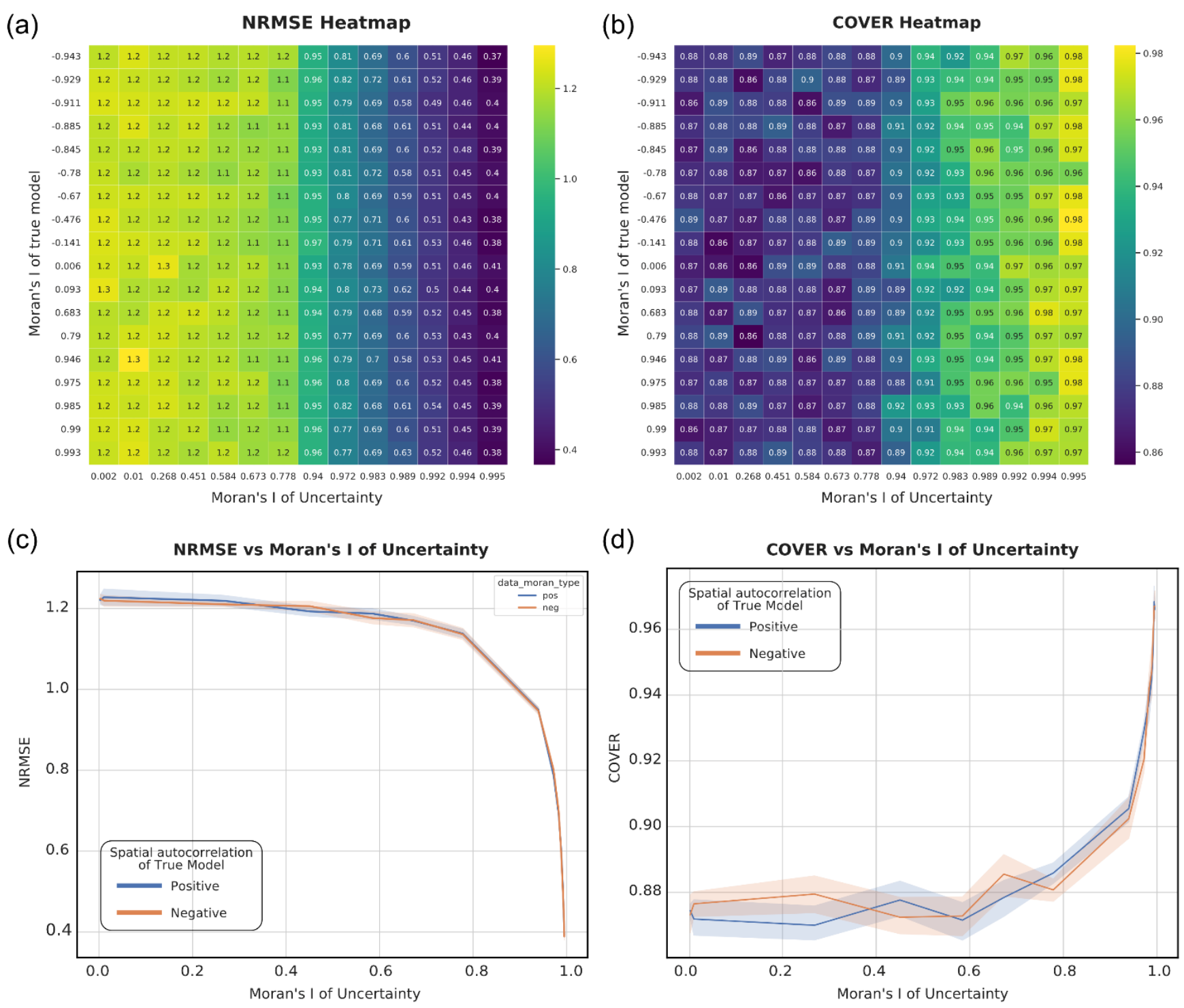

Figure 5 Summary of simulation results. Figures (a) and (b) demonstrate heatmap of two metrics (NRMSE, COVER) across Moran's I of uncertainty and Moran's I of true model. Figures (c) and (d) illustrate the relationship between Moran's I of uncertainty and these two metrics. Note: RMSE=root mean square error, COVER=coverage rate.

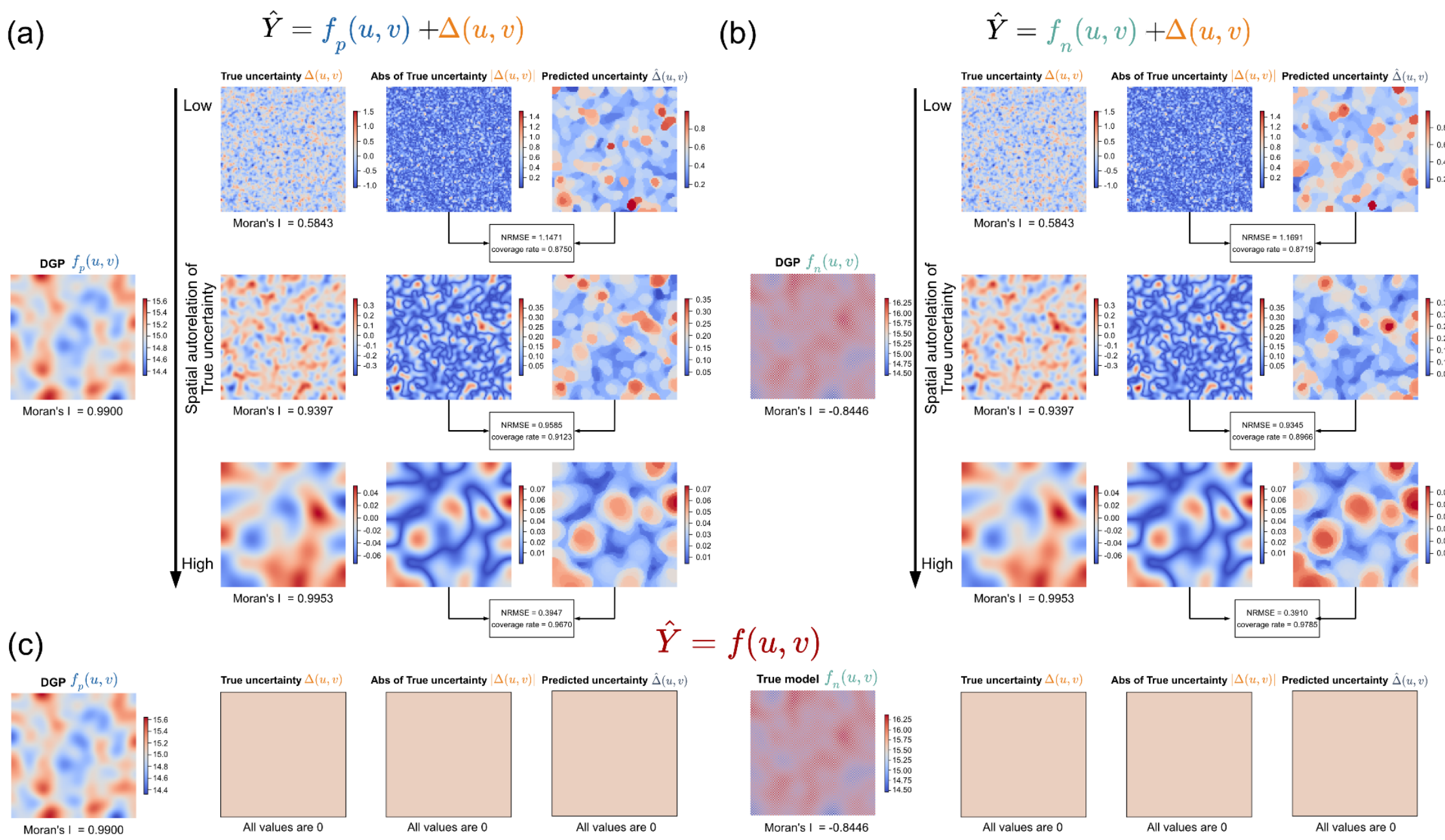

Figure 6 Results of three experiments. Figures (a) and (b) are the biased prediction models with simulated uncertainty, Figure (c) is the true prediction model without uncertainty.

**Experiment design for real-world datasets**

In this section, we describe the design, workflow, and objectives of our experiments, which are structured to assess the reliability and effectiveness of uncertainty estimates obtained from GeoCP across different spatial prediction frameworks. As shown in Figure 7, our experiments comprise three main components, focusing on spatial regression, spatial interpolation, and the impact of varied spatial representations on GeoAI prediction models.

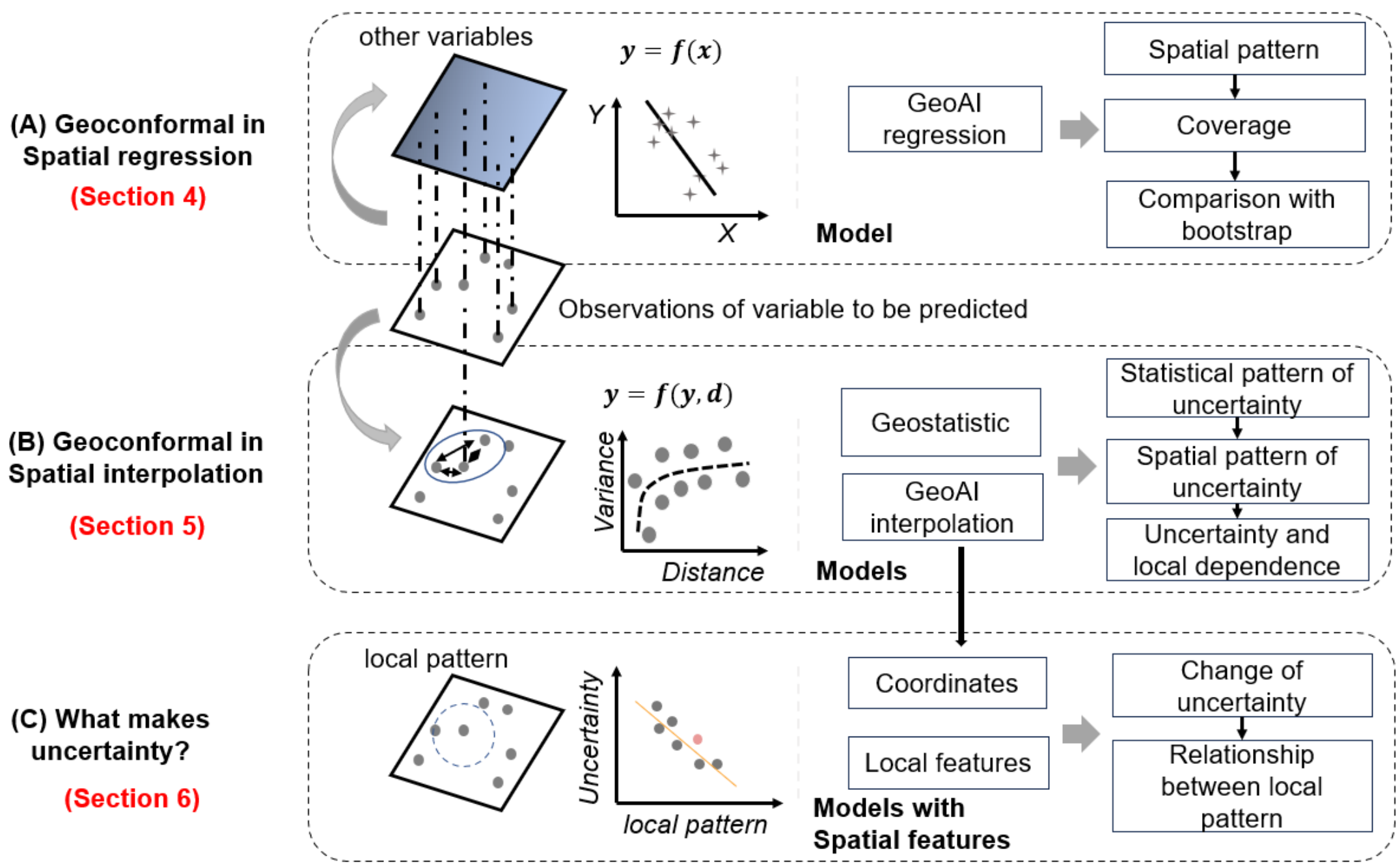

Figure 7 Experiment design of this study

First, in the spatial regression experiment (Figure 7a), we use GeoCP to understand how spatial features influence prediction uncertainty in a housing price model. We fitted the regression model on two types of features separately: aspatial features only and both spatial and aspatial features. Here, spatial features refer to georeferenced variables, such as coordinates; aspatial features refer to environmental and socioeconomic variables. Next, the 90%-level confidence intervals and corresponding coverage ratio were computed by GeoCP and Bootstrapping method, respectively. We then generated an error distribution for the regression model using bootstrapping, enabling a comparison of prediction uncertainty across different percentile levels. This experiment allowed us to validate the reliability of GeoCP uncertainty estimates and to explore how spatial characteristics affect uncertainty in spatial regression models.

Second, in the spatial interpolation experiment (Figure 7b), we focused on models that rely solely on the spatial distribution of the target variable, selecting Ordinary Kriging (OK) as a geostatistical model and Deep Geometric Spatial Interpolation (DGSI) as a GeoAI model. We explored the spatial characteristics of uncertainty in both models and examined its relationship with the spatial pattern of the target variable using Local Indicators of Spatial Association (LISA) and Moran's I index. This experiment allowed us to evaluate the reasonableness of uncertainty estimates in spatial interpolation models and to analyse how these estimates correlate with the spatial structure of the predicted variable.

Finally, we investigated how different representations of spatial information impact uncertainty in GeoAI-based prediction models (Figure 7c). Building on DGSI, we developed two new models that explicitly capture the local features of geographic variables. Using GeoCP, we compared the uncertainty of these models with that of the initial DGSI model, analysing how changes in uncertainty reflect the spatial distribution characteristics of the geographic variables. This component aimed to evaluate the sensitivity of GeoAI prediction models to different spatial representations, particularly how the enhancement of local features affects uncertainty estimates and aligns with the spatial patterns in the data.

## GeoCP in spatial regression

### *Experiment and data*

To assess the effect of spatial features in spatial prediction, the proposed GeoCP method is applied to a real-world dataset: housing sales prices and characteristics (e.g., building year, renovation year, number of floors, etc.) for Seattle. By adding or removing spatial features, we observed the difference in prediction uncertainty, thus identifying the impact of spatial features.

The dataset of housing prices was collected from GeoDa Lab (https://geodacenter.github.io/data-and-lab/KingCounty-HouseSales2015/), and it contains 21,613 observations with twenty-one related variables for Seattle and King County, Washington (May 2014 - 2015). We choose this dataset because of the high impact of geographic location on residential property prices. In this paper, we used a refined version of this dataset, which has eleven variables and only focuses on the Greater Seattle area. The dependent variable is the housing sale price (measured in $10,000), and the independent variables include eight aspatial features and two spatial features (UTM coordinates). The detailed description of this dataset is displayed in Table 1.

Table 1 Data summary of 2014-15 Seattle home sale

| | Explanatory Variable | Description |
|---|---|---|
| | price | Sale price |
| Aspatial | bathrooms | Number of bathrooms |
| Aspatial | sqft_liv | Size of living area in square feet |
| Aspatial | waterfront | ’1’ if the property has a waterfront, ’0’ if not |
| Aspatial | view | How good the view of the property was, from 0 to 4 |
| Aspatial | condition | Condition of the house, ranked from 1 to 5 |
| | grade | Construction quality, ranked from 1 to 13 |
| | yr_built | Year built |
| Spatial | UTM_X | House coordiate X under UTM coordinate system |
| | UTM_Y | House coordiate Y under UTM coordinate system |

***Results***

*The spatial distribution of the uncertainty*

In this part, we aimed to observe if the uncertainty measured by GeoCP is reasonable. We first trained two XGBoost Regressors on the Seattle housing price dataset. The first was fitted on aspatial features only, and the second was fitted on both aspatial and spatial features. Next, we generated geographic uncertainty with the GeoCP method, as shown in Figure 8a, b. For both models, the prediction uncertainty in the north is higher than that in the south. With location features (coordinates UTM X and UTM Y), the overall prediction uncertainty drops. The percentage of decrease in prediction uncertainty also varies over different regions, from 36.98% to 47.81% (see Figure 8c). From the geographic perspective, the prediction uncertainty in West Seattle, Renton, Kirkland, Redmond, Woodinville, and Cottage Lake declines the most, with values over 45%. This suggests that housing prices can be too overestimated or underestimated in these regions. Figure 8d,e illustrates the prediction errors for two XGBoost regressors. For XGBoost regressor with aspatial features only, the housing prices in Kirkland, Redmond, Woodinville, and Cottage Lake are underestimated; in Renton, they are overestimated; in West Seattle, both overestimation and underestimation happen. However, the prediction errors tend to be zero after introducing coordinates as features, implying that the housing prices are highly influenced by their geographic locations. We also mapped the change rate of prediction errors between two XGBoost Regressors (see Figure 8f). The prediction errors at most locations show a declining trend when both aspatial and spatial features are inputted. The high overlap between GeoCP uncertainty and prediction error proves that GeoCP uncertainty successfully reflects the effect of spatial features. This makes our proposed GeoCP a promising method to explain the spatial effect.

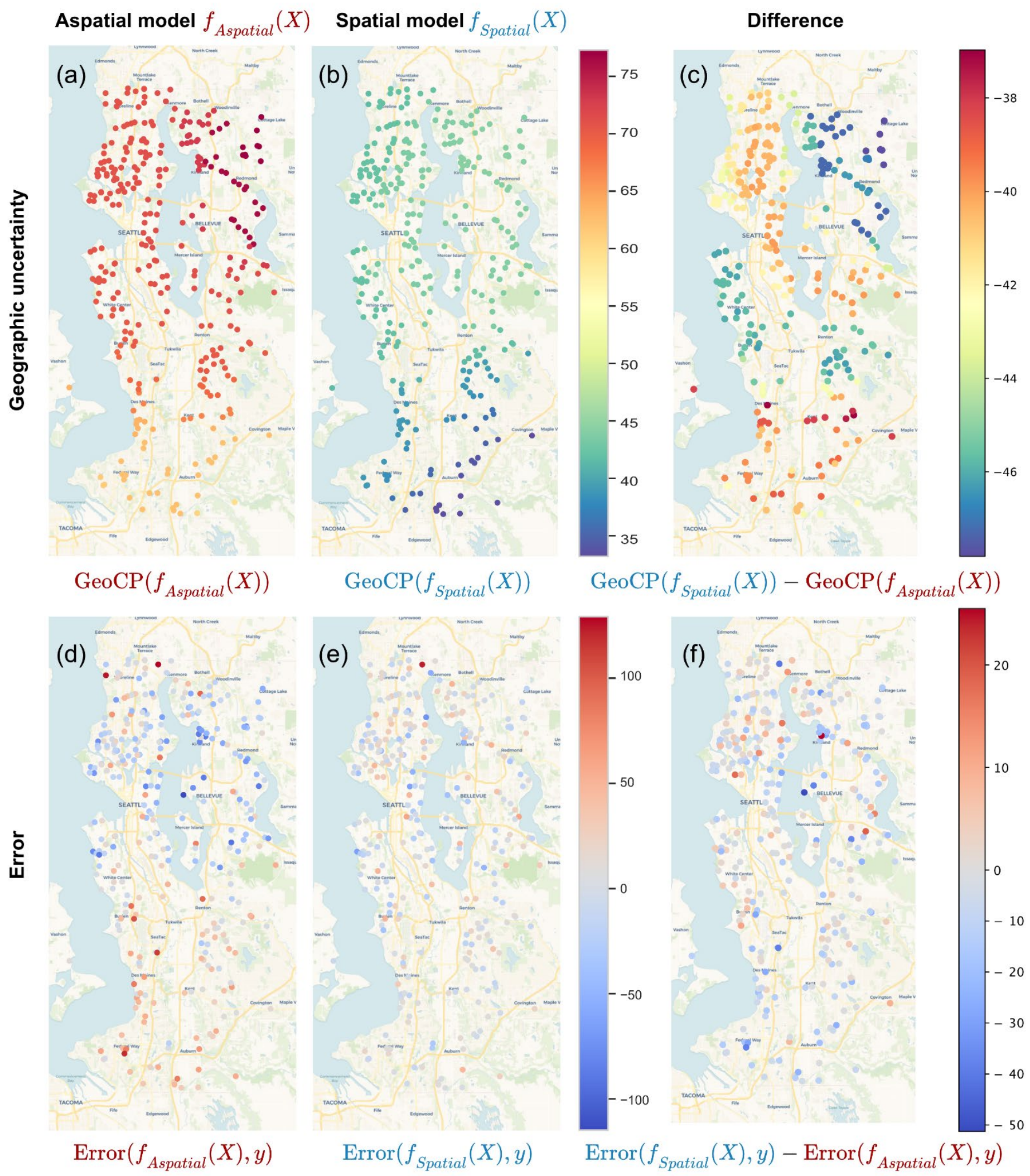

Figure 8 Geographic uncertainty and error of aspatial and spatial models on Seattle house price dataset. Figures (a) and (b) demonstrate the geographic uncertainty for aspatial and spatial models, Fig (c) shows the difference of uncertainty between two models. Figures (d) and (e) illustrate the errors for aspatial and spatial models, Fig (f) displays the difference of errors between two models.

*Evaluation of coverage ratio*

To further study the reliability of the estimated uncertainty, we used coverage ratio as an indicator, which representing the proportion of test samples whose predicted values fall within the prediction interval. In this paper, we compared the 90%-level confidence interval measured by GeoCP and Bootstrapping method. The uncertainty interval predicted by GeoCP is a prediction interval, as discussed in Appendix A. To compare the results from GeoCP and the Bootstrapping method, we calculated the prediction interval for the Bootstrapping method following the process outlined in Appendix B.

The procedure of comparison between GeoCP and Bootstrapping is described below. First, a host of new Bootstrap datasets are created by repeatedly resampling the housing price dataset with replacement. Here, we resampled the dataset for 2000 times to create 2000 Bootstrap datasets. Then, XGBoost regressors were fitted on 2000 bootstrap datasets. For each location, 2000 predicted values were generated. By taking 95% and 5% percentiles of these 2000 values, the confidence interval for each location can be obtained. As shown in Figure 9a, the GeoCP offers a much higher coverage ratio (about 93.6%) than bootstrapping. This value slightly overtakes the given confidence level, suggesting the confidence interval computed by GeoCP is reliable and suitable for interpreting uncertainty. However, the maximum coverage ratio of the Bootstrapping method is about 81.00%. In addition, the coverage ratio for the Bootstrapping method increases as the resampling times are added, from 62.67% with 10 times to 80.00% with 500 times (see Figure 9b). Ultimately, the coverage ratio will fluctuate between 79.33% and 81.00 % even if the resampling times continue to grow. This suggests that using a Bootstrapping confidence interval may be misleading. It is noted that the time consumed for calculating the bootstrapping confidence level (2000 resampling times) on a PC with an i7-6700 CPU and 16GB RAM is 524.76s. This time is much longer than the GeoConformal prediction (0.4254s) because the model only needs to be fitted once.

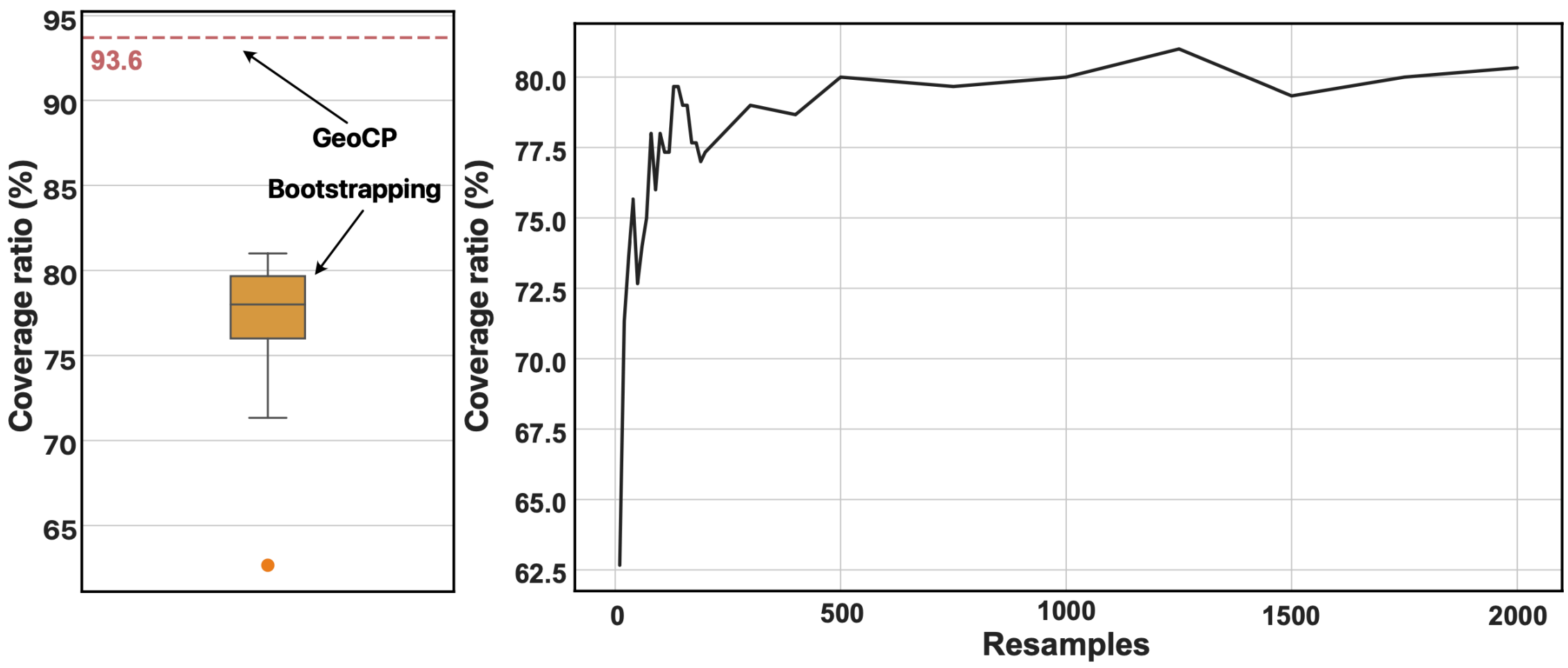

Figure 9 Coverage rate for GeoCP and Bootstrapping

Although we have proved the superiority of our proposed method in coverage ratio, we still need to clarify the relationship between GeoCP uncertainty and actual error distribution. The estimated errors for 2000 Bootstrap datasets were computed. We sorted these errors and took the different percentiles directly (30%, 50%, 70%, 90%) at each location. The relationship between predicted errors and GeoCP uncertainty at different percentile levels is plotted in Figure 10. It is obvious that at any percentile level, the prediction error shows high linearity with uncertainty, with correlation coefficients of 0.9788, 0.9222, 0.8760, and 0.9385, respectively. The result implies that GeoCP uncertainty is capable of reflecting actual error distribution.

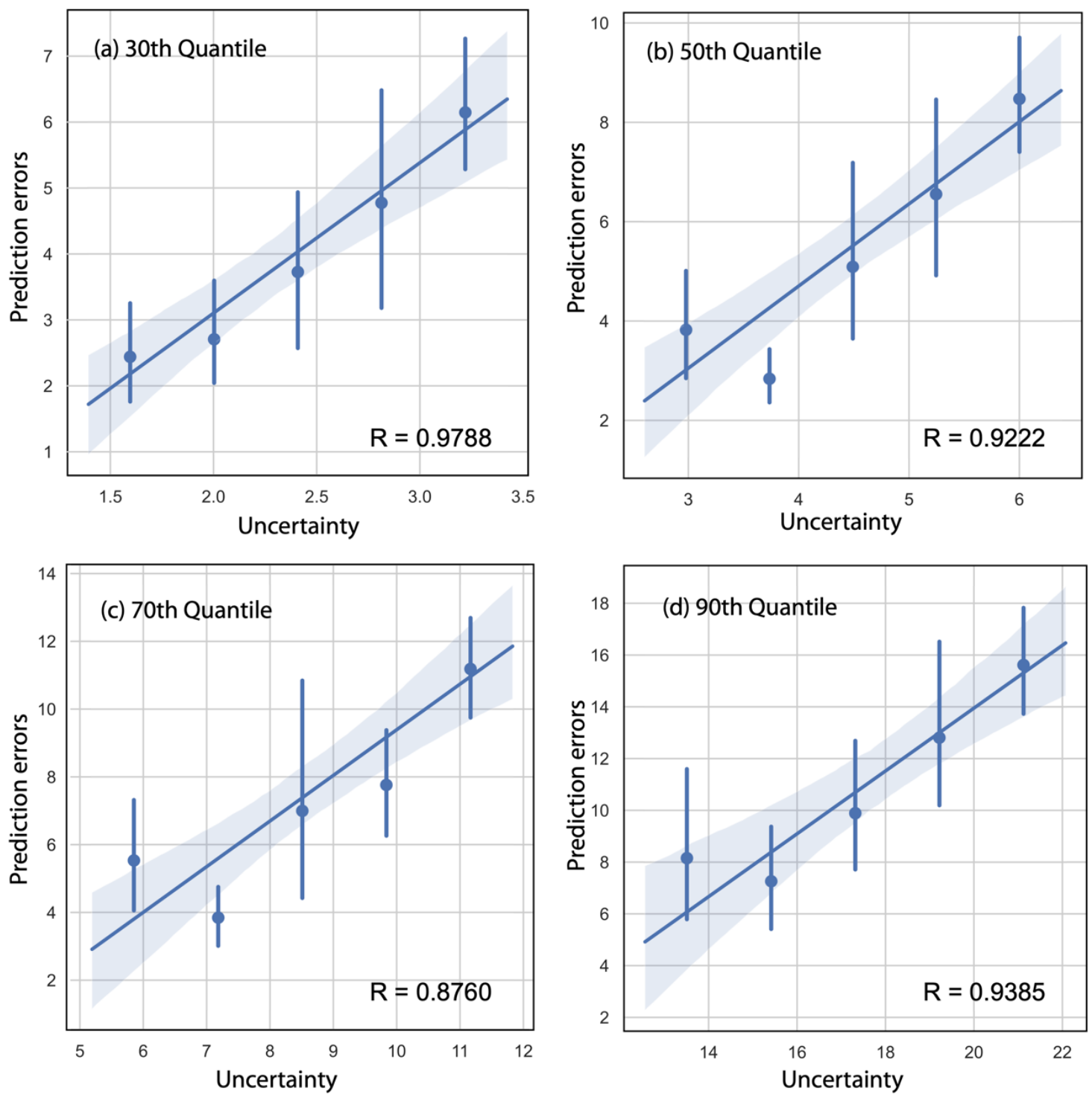

Figure 10 Prediction error of housing price with aspatial and spatial features

The strong coverage guarantee of GeoCP ensures that the uncertainty estimated is reliable and comparable across different models and provides trustworthy interpretation for data. The high linear relationship between the uncertainty and prediction errors suggests that both methods capture the model's error distribution effectively. This mutual validation also strengthens confidence in the computed uncertainty. The GeoCP uncertainty has an advantage over the Bootstrapping method.

## GeoCP in spatial interpolation

In this section, we tried to apply the GeoCP to another classical task in GIS: spatial interpolation. Spatial interpolation predicts the value at an unobserved location by directly modelling the spatial dependence (such as the semivariogram in Kriging methods) instead of relying on the explanatory variables. The generated geographic uncertainty is more directly related to the spatial structure and values of the predicted geographic variable. As a result, spatial interpolation is a very suitable scenario for us to explore in depth the ability of different types of interpolation methods to capture spatial patterns. In this section, Ordinary Kriging (OK) and a deep-learning-based interpolation method were chosen. GeoCP uncertainty of OK and Kriging variance were compared to verify the reliability of GeoCP uncertainty. Meanwhile, the GeoCP uncertainty of the two models was compared to analyse the difference in modelling spatial structure between these two models.

***Experiment and data***

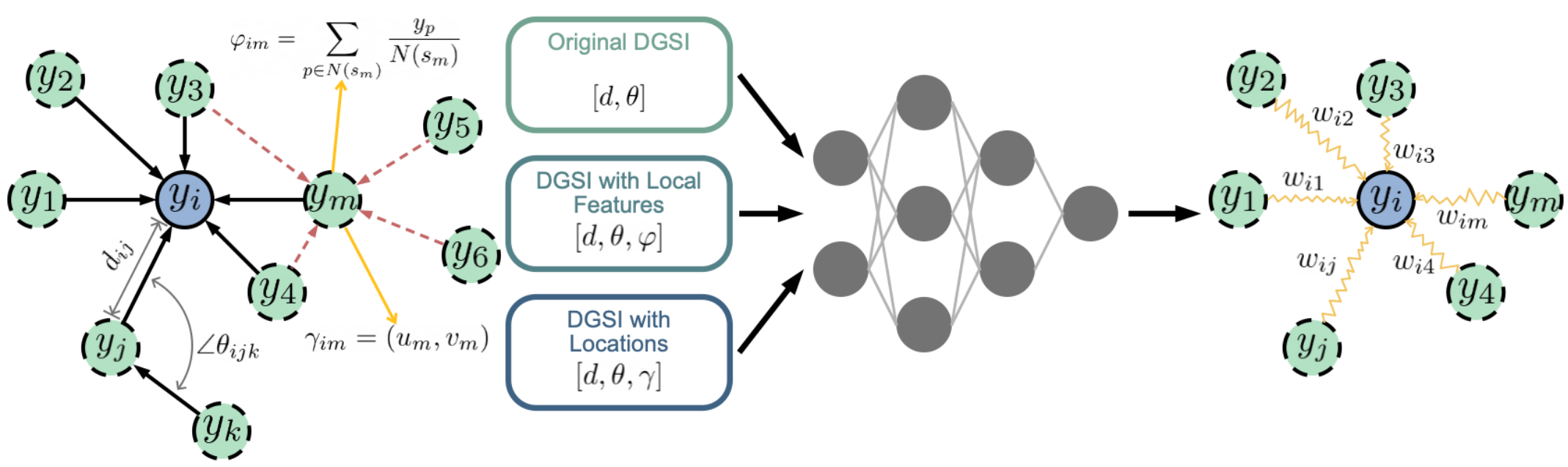

Figure 11 DGSI and its two improvements (with local feature, and with location information)

We selected two spatial interpolation models: one is a traditional geostatistical model, Ordinary Kriging (OK), and another is a GeoAI model, Deep Geometric Spatial Interpolation (DGSI) (Zhang et al. 2022). Ordinary Kriging is one of the most widely used methods for spatial interpolation. It estimates values at unknown locations based on nearby known points. Considering similar values tend to be near each other, close points are given higher weights. With local second-order stationarity assumption, ordinary kriging implicitly measures the mean in a moving neighborhood (Wackernagel 1995). However, this assumption may not hold in many situations. DGSI is a deep learning-based framework for spatial interpolation. Traditional spatial interpolation methods usually rely on predefined spatial distributions or kernel functions. DGSI addresses this limitation by learning spatial relationships with multilayer neural networks. This model incorporates both distance and orientation information between data points to improve accuracy, considering both spatial autocorrelation (nearby locations tend to be similar) and heterogeneity (each location can have unique attributes). Figure 11 illustrates the DGSI and its two variants, which will be used in Section 6. The original DGSI employs merely distance and orientation information as input. On the basis of the original DGSI, DGSI with local features adds the average of $k$-nearest neighbors' target values $\{y_i\}_{i\in\mathcal{N}_k}$, while DGSI with locations directly adds the coordinate values of the $k$-nearest neighbors $\{u_i, v_i\}_{i\in\mathcal{N}_k}$, where $\mathcal{N}_k$ denotes the $k$ points with smallest distances to the point to predict.

The evaluation involves two groups of experiments. One is to measure the prediction uncertainty and verify its reliability. First, we demonstrated that geographic uncertainty computed by GeoCP works as expected. Then, we explored the statistical and spatial patterns of prediction uncertainty for both OK and DGSI. Next, we verified the ability of geographic uncertainty to capture spatial patterns at local and global scales. The other experiment is to investigate the influences of spatial dependencies on the prediction uncertainty and deepen the understanding of what makes the uncertainty of spatial prediction. First, we improved the DGSI with two different spatial features: local features and location information. Second, we compared the difference in prediction uncertainty when applying different spatial features. Finally, we computed the statistical relationship between local spatial dependence and change in the prediction uncertainty influenced by spatial features.

Because of strong spatial autocorrelation and wide availability, predicting temperature at unobserved locations has been an important application for spatial interpolation. We evaluated the performance of the GeoCP method in spatial interpolation on a 90-day ambient temperature dataset collected from Weather Underground. This dataset covers the region of Los Angeles County from 1st January 2019 to 31st March 2019. For each day, there are 90 sample points, with an 80%/10%/10% split for training, calibration, and test sets adopted.

## ***Results***

### *Measured geographic uncertainty*

We randomly selected one day (i.e., the 46th day) as an example to visualize the results. DGSI was applied to predict the temperature, and the uncertainty—specifically, the length of the prediction interval—was measured using both the original conformal prediction and GeoCP methods. The measured uncertainty results are illustrated in Figure 12.

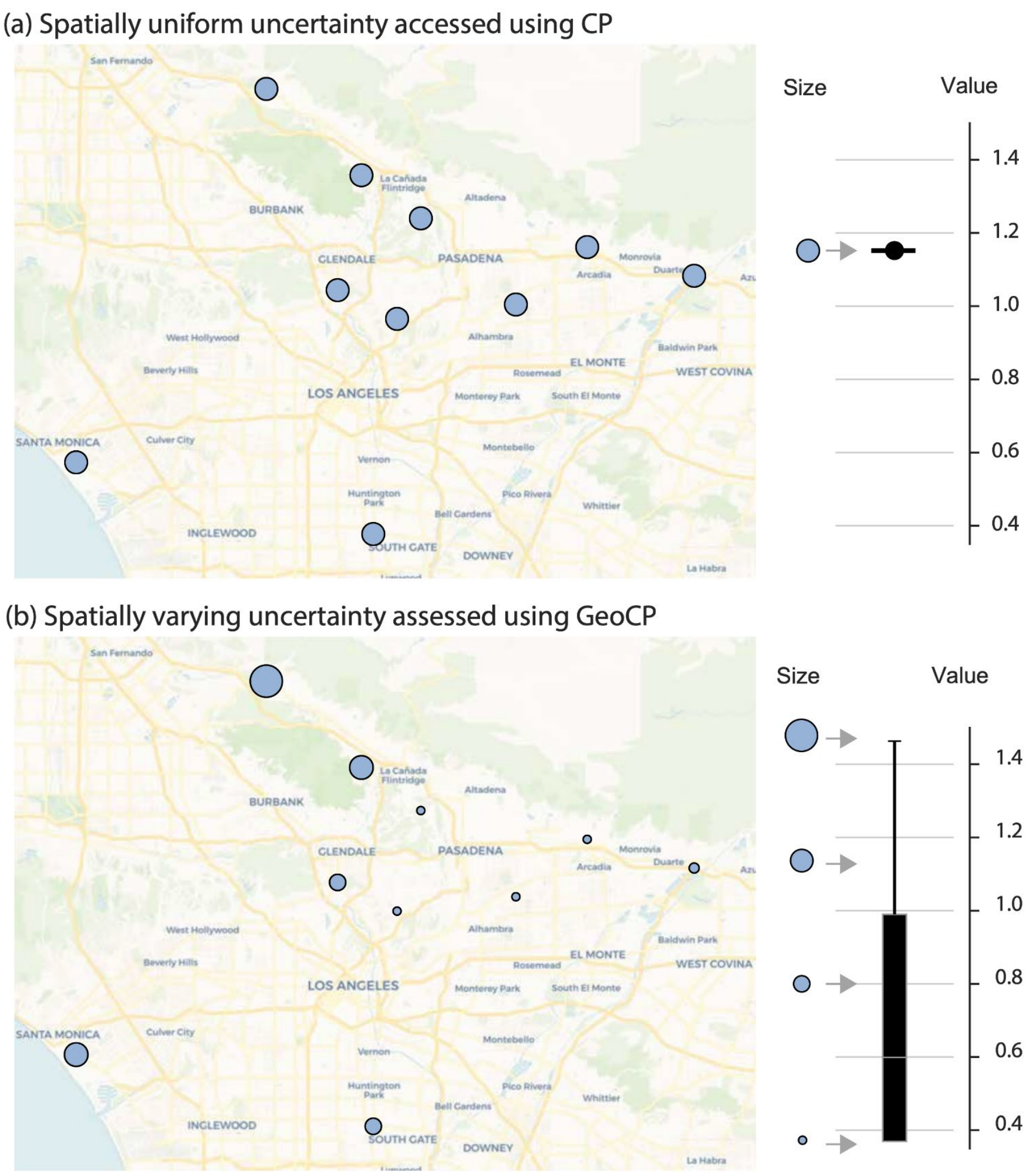

Figure 12 Uncertainty measured by conformal prediction

In the original conformal prediction, every nonconformity score from the calibration set is equally weighted, so every location has the same length of prediction interval, 1.15 (see boxplot in Figure 12a). By applying geographic weights dynamically, the uncertainty changes over different geographic locations, from 0.3 to 1.5 (see boxplot in Figure 8b). The sizes of points stand for the prediction difficulty for the model. For example, in the map in Figure 12b, the northernmost point has the biggest size in comparison with other points, suggesting the temperature at this location is hard for the model to predict. It is worth noting that because the original conformal prediction can only offer a fixed prediction interval, the sizes of points in the map in Figure 8a are just the same. The spatially varying characteristics of GeoCP enable us to conduct a place-based uncertainty analysis.

*Statistical pattern of uncertainty*

To have an overview of the uncertainty for different models, we used original conformal prediction to calculate the uncertainty of each day. Here, we select three OK models with different variogram models: exponential, linear, and Gaussian models. For simplicity, we term these three kinds of OK as exponential OK, linear OK, and Gaussian OK. The choice of variogram model poses a great impact on the prediction error and uncertainty for OK (see boxplot in Figure 13). Gaussian OK performs worst, with the largest RMSE (root mean square error) and uncertainty, suggesting that the Gaussian model may fail to capture the spatial structure. In terms of RMSE (see Figure 13a), DGSI obviously outperforms exponential OK and linear OK, with an average value of less than 1. However, the prediction uncertainty of DGSI is larger than that of exponential OK and linear OK (see Figure 13b). In general, exponential OK, and linear OK have very similar average prediction with DGSI, all around 1. Since exponential OK outperforms other OK models, we will continue to use exponential OK to compare with DGSI in the following sections.

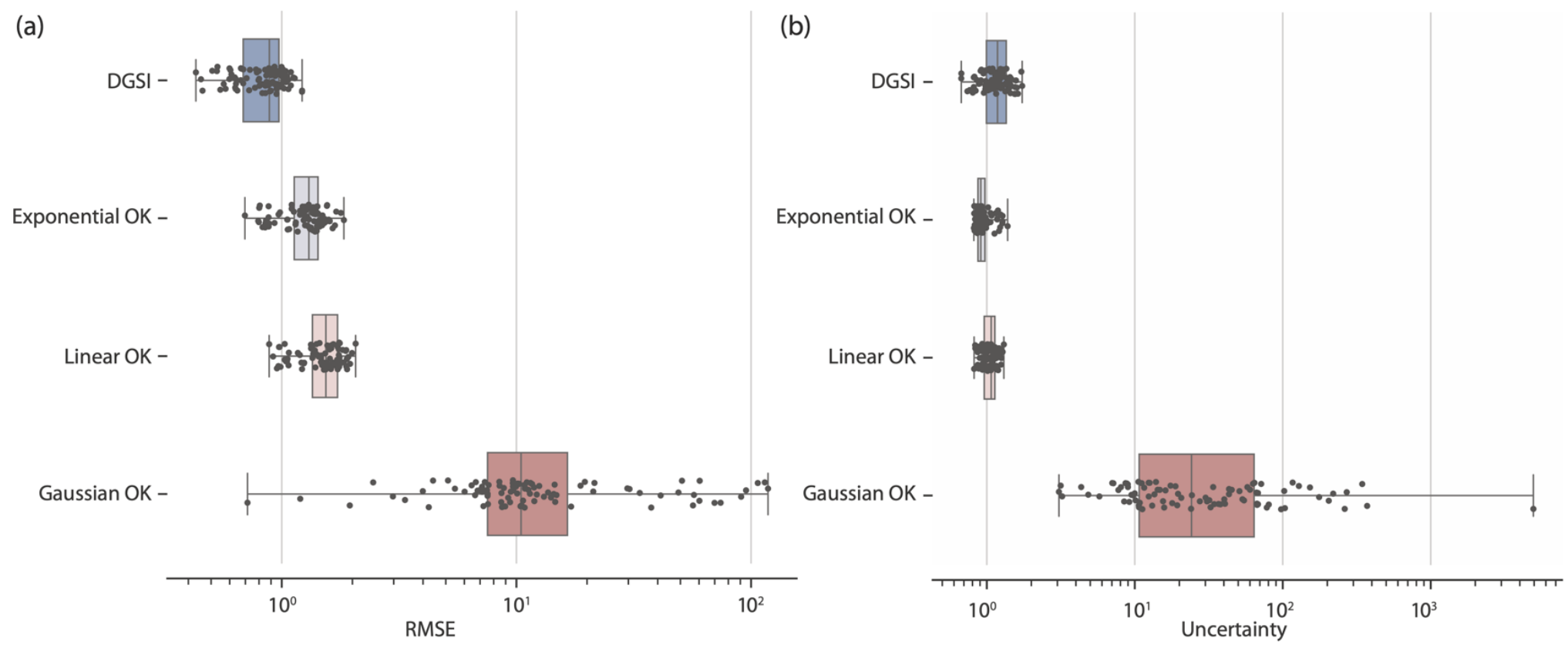

Figure 13 Prediction error for OK and DGSI

*Spatial pattern of uncertainty*

After analysing uncertainty for different models from a statistical perspective, we then applied GeoCP to calculate the geographic distribution of uncertainty for all 90 days. Moran's I was used to measure the spatial autocorrelation of geographic uncertainty, as demonstrated in Figure 14. The geographic uncertainty of DGSI shows a strong spatial pattern, with most Moran's I values above 0.1. However, the geographic uncertainty of OK has no significant spatial autocorrelation, with most Moran's I values close to 0. We also calculated the kriging variance for every day. Kriging variance refers to the uncertainty in predictions made using the OK method. The kriging variance also has a low spatial autocorrelation. All the Moran's I of kriging variance are around 0.01. The reason that the spatial autocorrelation of the uncertainty and variance for OK is low may be that the OK method only considers the distance between observed and unobserved points, so it cannot learn the actual spatial structure from the dataset. From a statistical perspective, DGSI performs well both in error and uncertainty. Nevertheless, DGSI's uncertainties show higher spatial patterns when extending to spatial dimensions, which may lead to geographic bias.

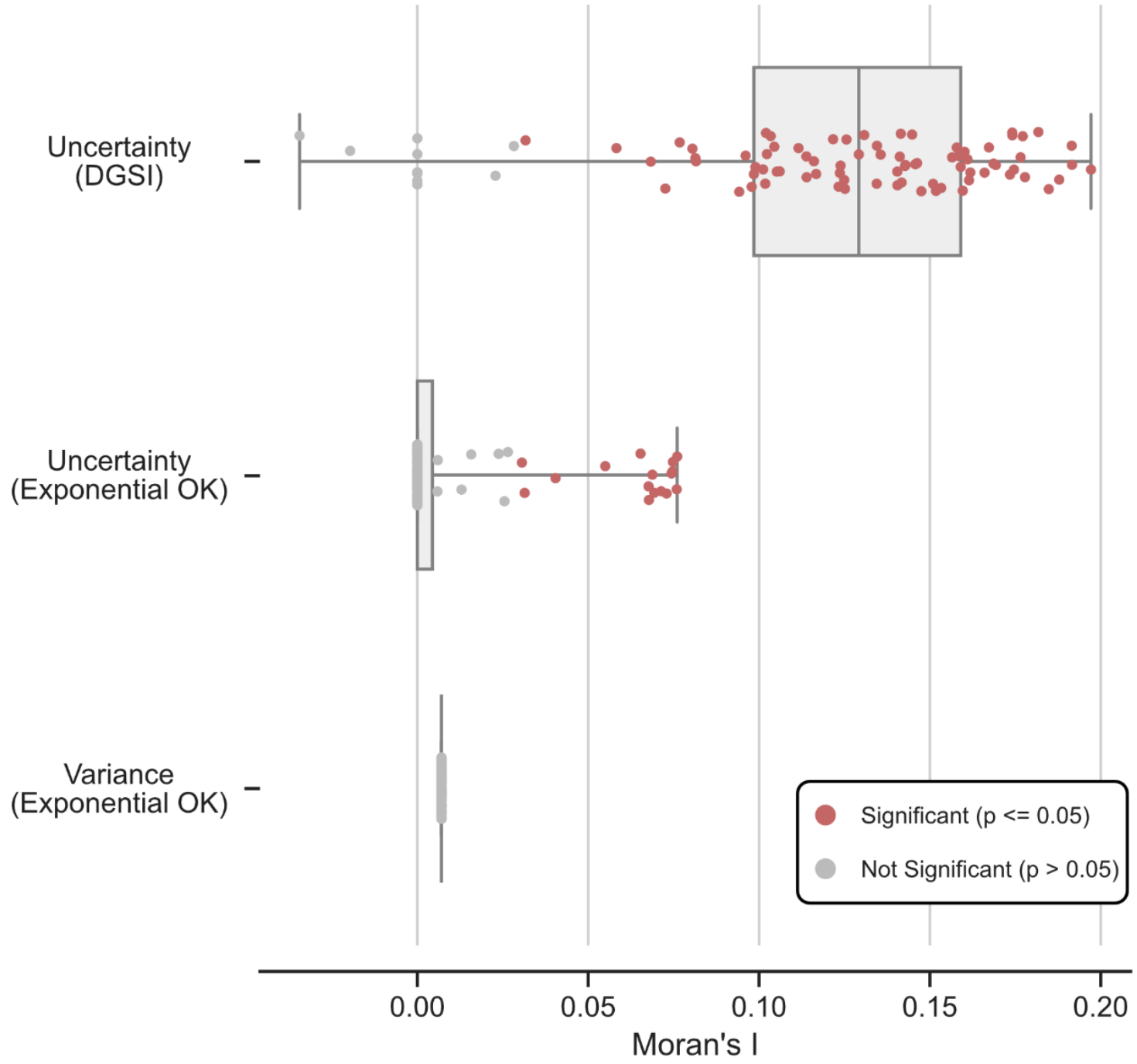

Figure 14 Moran's I of geographic uncertainty for exponential OK and DGSI. OK variance is also included to make a comparison with geographic uncertainty.

*Geographic uncertainty unveils local spatial dependence*

We further took local spatial dependence into consideration to analyse the geographic uncertainty. As illustrated in Figure 15, we first computed the correlation coefficient between the local spatial dependence, such as local Moran's I, and geographic uncertainty, then explored its relationship with the global spatial dependence, such as Moran's I. The Moran's I and the correlation between uncertainty and local Moran's I show significant linearity; that is, with a stronger spatial pattern (higher Moran's I), the prediction uncertainty is affected more positively by local spatial pattern and vice versa. It is noted that this linear relationship for DGSI (Pearson correlation coefficient = 0.6091) is much stronger than that for exponential OK (Pearson correlation coefficient = 0.2735), implying that DGSI can better capture spatial dependence in the geospatial data. This amazing ability of GeoCP uncertainty allows us to have a deeper understanding of different geospatial models and how they learn the spatial structure of data. In this way, we can design more responsible and trustworthy geospatial models.

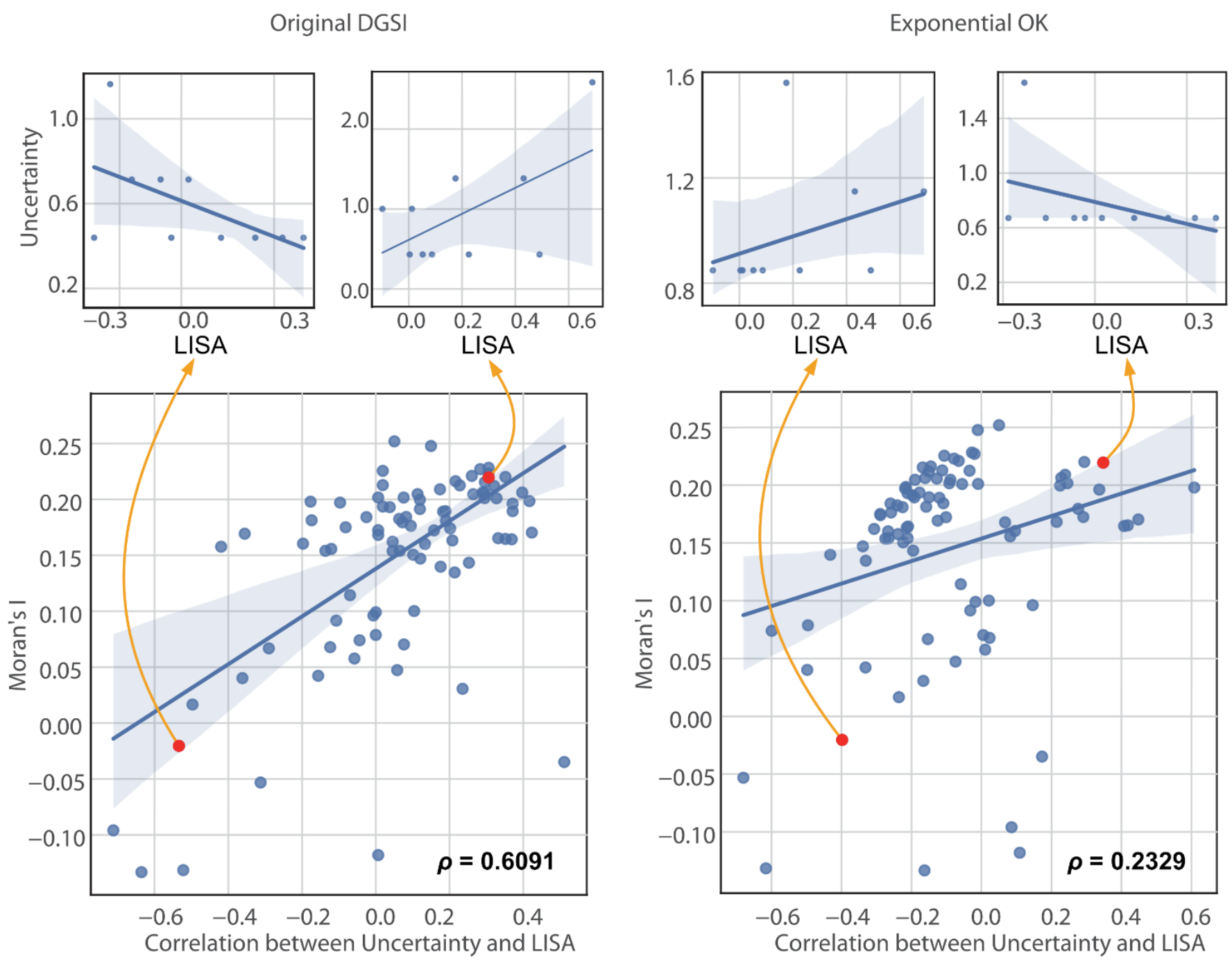

Figure 15 Global spatial dependence versus correlation between geographic uncertainty and local spatial dependence.

**Understanding the relationship between prediction Uncertainty and spatial effects**

To further understand the uncertainty of spatial prediction, we integrated some spatial features into DGSI (see Figure 11) to observe how they change the uncertainty of spatial prediction. In this paper, the mean value of neighbors (local feature) and coordinates (location information) were used as spatial features.

***Spatial features lead to a decrease in uncertainty***

By adding spatial features, we attempted to see their influences on the uncertainty from both statistical and spatial perspectives. Both types of spatial features can reduce an overall uncertainty by 0.05 (see Figure 16a). In addition, the uncertainty decreases more when more local features are added, suggesting that local features can help the DGSI better learn the spatial dependence inside the geospatial dataset than the location information, thus giving the model more confidence in prediction. The introduction of spatial features also helps reduce Moran's I of prediction uncertainty. As shown in Figure 16b, the Moran's I of prediction uncertainty of three DGSI models all concentrates on the areas around 0.0 and 0.15. However, the peaks of DGSI with Location and DGSI with Local features are on the left of that of original DGSI, that is, Moran's I of prediction uncertainty for DGSI models with spatial features is smaller in most cases. This means that the presence of spatial features can reduce the spatial autocorrelation of prediction uncertainty. Apparently, apart from giving more confidence in prediction, introducing more spatial features also reduces the spatial pattern of the uncertainty. This sheds light on how to design a responsible model with less inequality and more fairness. In a nutshell, by introducing more spatial features, the overall geographic uncertainty and its spatial autocorrelation decrease, thus making the uncertainty distribution more evenly across different regions or strata.

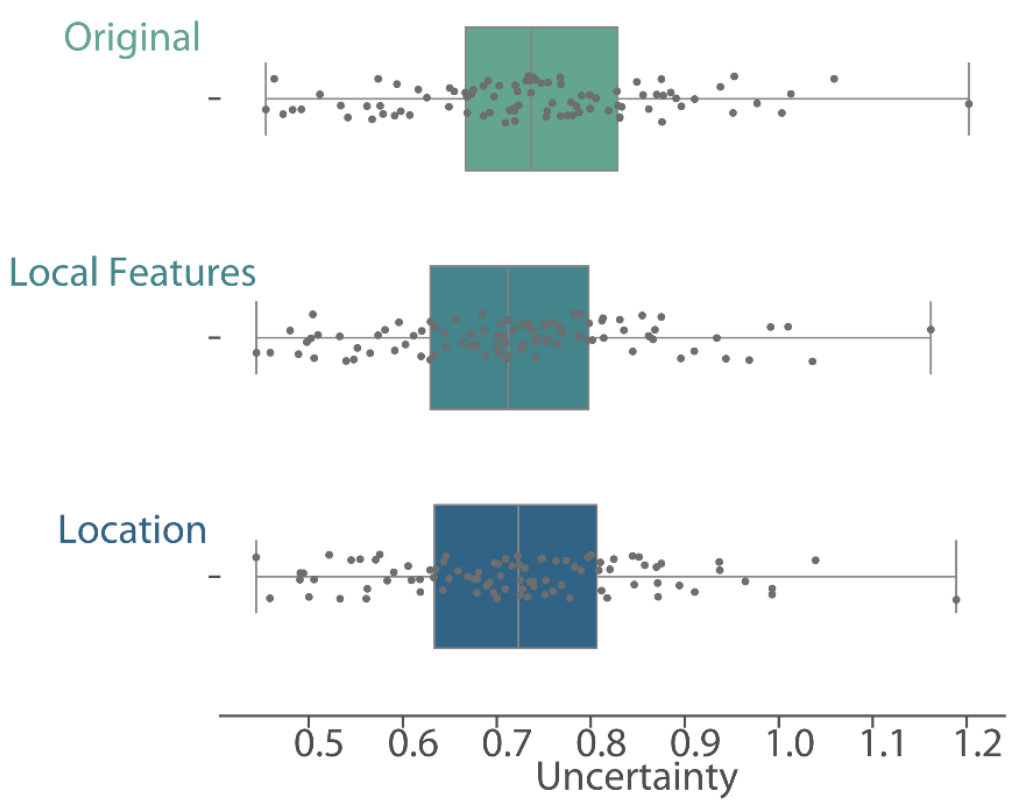

(a) Distribution of geographic uncertainty for original DGSI and its two improved versions.

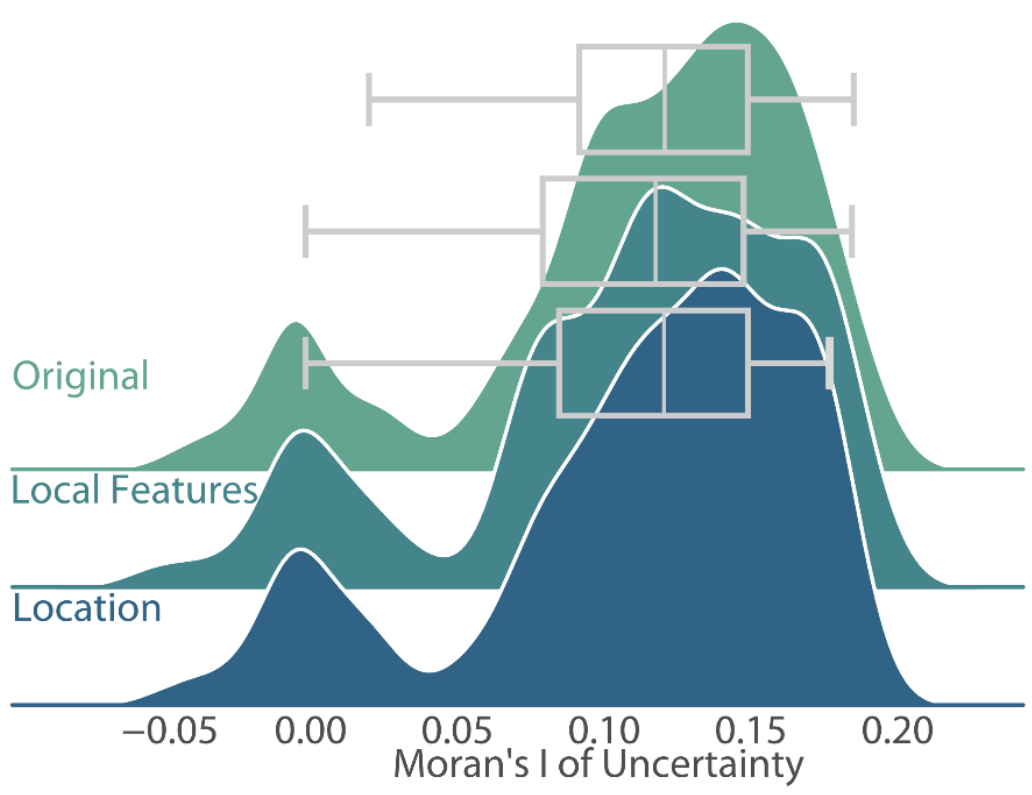

(b) Distribution of Moran's I of geographic uncertainty for original DGSI and its two improved versions. Y-axis represents the density for a specific Moran's I value.

Figure 16 Geographic uncertainty and Moran's I of uncertainty for different DGSI models (original, with local feature, and with location information)

***Spatial dependence produces uncertainty***

To evaluate how spatial dependence influences uncertainty, we computed the correlation between local spatial dependence and change in uncertainty due to different spatial features. In the previous section, we demonstrated that explicitly modelling spatial structure can effectively reduce prediction uncertainty. As shown in Figure 17a, the magnitude of this uncertainty reduction is strongly associated with the strength of local spatial dependence. This indicates that the primary mechanism behind the reduction lies in the model's enhanced ability to capture spatial dependence. Accordingly, when spatial structure is not explicitly or adequately modelled, the presence of unaccounted spatial dependence may become a key source of prediction uncertainty. The relationship between local spatial dependence and uncertainty changes suggests that researchers should pay more attention to places with high local spatial dependence, where a higher uncertainty may be generated.

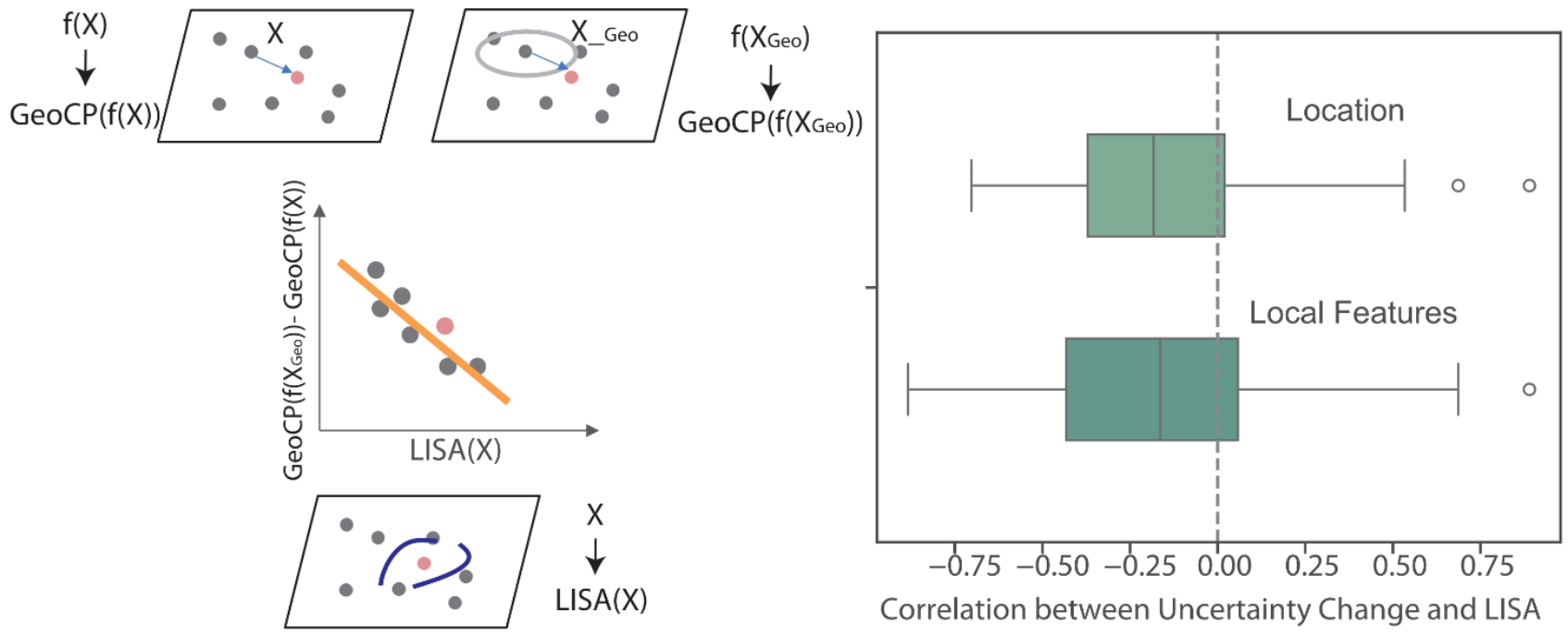

(a) Correlation between geographic uncertainty change and local spatial dependence. Left part shows the calculation process of correlation coefficients. Right part shows the distribution of correlation coefficients for two improved versions of DGSI.

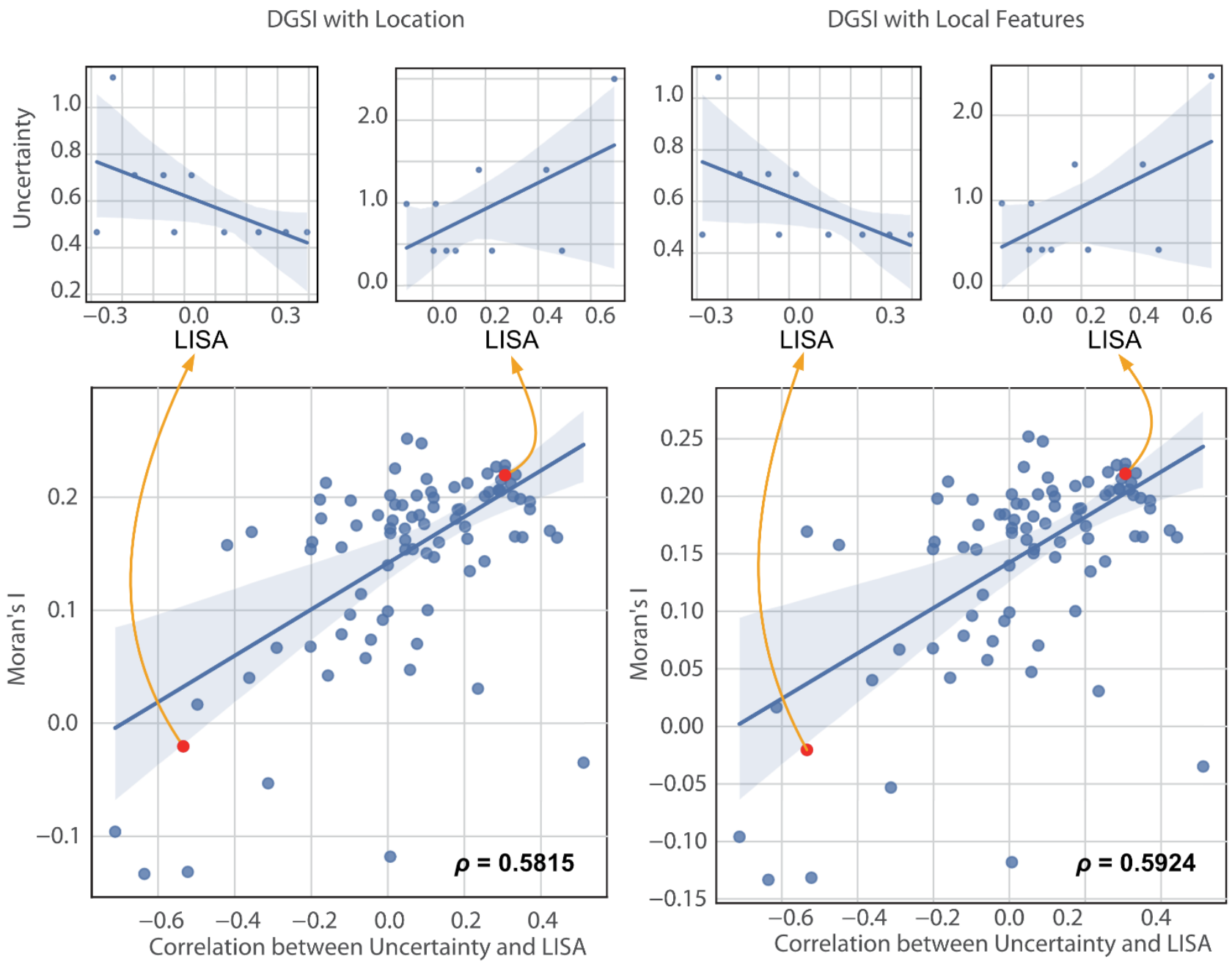

(b) Global spatial dependence versus correlation between uncertainty and local spatial dependence for two improved versions of DGSI.

Figure 17 Exploratory results for two improved versions of DGSI.

## Discussion

### *Significance of GeoCP*

Given its performance in spatial regression and spatial interpolation, GeoCP proves a powerful framework for measuring the prediction uncertainty of different geospatial models, thus helping us to better understand what is causing geographic uncertainty or bias. The significance of GeoCP can be summarized in three aspects. First, the model-agnostic feature of GeoCP makes it possible to compare various geospatial models, from traditional algorithms such as geographically weighted regression (GWR) and Ordinary Kriging (OK) to complicated methods such as XGBoost and deep neural network (DNN) without modifying their internal structures. Second, GeoCP allows an intuitive interpretation of uncertainty by directly outputting prediction intervals or prediction sets rather than single-point estimations. For instance, the prediction in areas not well represented in the training set is difficult, so the corresponding uncertainty will be high. What is most important is that conformal uncertainty can have a clear physical meaning depending on the variable to be predicted. With the interpretation power of GeoCP, we may explore the causal relationship between uncertainty and diverse factors, such as uneven sampling of geospatial data and a special design in a neural network. Third, by introducing geographic weighting, GeoCP is robust to distribution shift. Because conformal prediction merely assumes exchangeability and relies on weak distributional assumptions, GeoCP is able to adapt automatically according to the changes in model performance. For example, if a regression model starts to perform well in a new location, the prediction intervals will be shortened to reflect reduced uncertainty.

***The source of uncertainty in spatial prediction***

From the experiments above, we found a strong relationship between uncertainty and spatial features as well as spatial patterns. Here, we would like to summarize some insights about the source of uncertainty in spatial prediction. In spatial regression, the introduction of spatial features into explanatory variables helps reduce prediction error and uncertainty. Areas where housing prices are much overestimated or underestimated are recognized by visualizing the uncertainty change (see Figure 8c). This makes our proposed method a powerful tool to identify the performance of different models across different regions. In our case, the housing sale prices are highly influenced by their locations, indicating accordingly that the uncertainty is also highly affected by the spatial features. In spatial interpolation, the conformal uncertainty for the GeoAI model and ordinary kriging is close to each other (see Figure 13). However, when extending to the geospatial context, the GeoCP uncertainty of the GeoAI model tends to show a stronger spatial pattern than that of ordinary kriging (see Figure 14). This phenomenon indicates that compared with the statistics-based model (such as Kriging), the GeoAI model fails to capture spatial variability. As a result, it is essential to explicitly represent spatial information when designing a GeoAI model. Moreover, we discovered that as global spatial dependence strengthens, the uncertainty is more positively affected by local spatial dependence (see Figure 15). Joining two different spatial information representations in the experiments reveals that the uncertainty can be reduced significantly by allowing the GeoAI model to learn more spatial knowledge. This implies that the uncertainty may come from spatial dependence or spatial structure. For a GeoAI model, learning local features can help reduce uncertainty more than just learning location information (coordinates) (see Figure 16). We also proved that the stronger the local spatial dependence, the more effective the explicit representation of spatial information is in reducing uncertainty (see Figure 17).

***Future work***

In this paper, we conducted preliminary experiments to investigate the effects of geospatial features on geographic uncertainty, laying the groundwork for exploring the "why" problem of geographic uncertainty. Moving forward, the uncertainty quantification provided by GeoCP can be incorporated into GeoAI models, promoting the development of responsible geospatial intelligence. The design of GeoAI models should aim to not only decreasing the prediction uncertainty but also reducing the geographic bias. In the previous research (Wu et al., 2024), the geographic bias is usually computed based on the model errors. Nevertheless, the error varies within a certainty range for each location, which makes it improper to represent the realistic situation. As a result, the calculated geographic bias may not be accurate. In GeoCP, the uncertainty at each location reflects the range of the prediction error distribution, reflecting the maximum possible error. As a result, GeoCP uncertainty can be a reasonable way to study geographic bias.

Another promising direction is to compare and integrate different uncertainty quantification frameworks with GeoCP. Among them, Bayesian-based methods are widely used and well-established, with conceptual foundations that are fundamentally different from those of conformal prediction. Conformal prediction is rooted in the frequentist paradigm and provides distribution-free, finite-sample coverage guarantees for unseen data, without requiring a probabilistic model of the data-generating process. In contrast, Bayesian methods adopt a probabilistic framework, modeling uncertainty through a full posterior distribution over predictions conditioned on both the data and prior beliefs. In summary, while Bayesian approaches offer a rich, model-based characterization of uncertainty, conformal prediction ensures valid coverage levels under minimal assumptions. These complementary strengths open exciting opportunities for hybrid frameworks. For instance, Bayesian techniques could be used to enhance the underlying predictive model within the GeoCP framework, allowing more expressive modeling of epistemic uncertainty, while GeoCP contributes distribution-free coverage guarantees. Such integration could lead to more robust, scalable, and interpretable uncertainty quantification in spatial prediction tasks.

## Conclusion

This paper introduces a powerful model-agnostic framework named GeoConformal Prediction (GeoCP) for measuring geographic uncertainty. GeoCP extends the conformal prediction method by integrating geographic weighting. The GeoCP uncertainty value reflects the prediction interval or set for each location, providing an intuitive tool for researchers to study uncertainty across different geospatial models. The proposed method is applied to two tasks, namely spatial regression and spatial interpolation. In the spatial regression, GeoCP is proved to be reasonable and reliable because of its high coverage ratio and significant linear relationship with actual error distribution. In spatial interpolation, GeoCP reveals that geographic uncertainty is highly related to local spatial dependence. Further experiments with two spatial features (local features and location information) show that the introduction of spatial features can reduce uncertainty itself as well as its spatial autocorrelation. The proposed GeoCP method can further be used in measuring geographic bias, thus building responsible and trustworthy geospatial models and possibly alleviating social inequality.

## Data and codes availability statement

The data and codes that support the findings of the present study are available on Github at https://github.com/pengtum/GeoCP/.

## Appendix

### *A: The difference between confidence interval and prediction interval*

Prediction interval (PI) quantifies the uncertainty in estimating the values of new observations. Confidence interval (CI), in contrast, captures the sampling uncertainty, that is. variability in the estimate of a population parameter, such as the mean value. Given a 95% PI of $[lb_p, ub_p]$ and a 95% CI of $[lb_c, ub_c]$, a 95% PI suggests that the actual value of an unobserved point is 95% sure to fall in the range $[lb_p, ub_p]$ while a 95% CI implicates that the mean value of the data is 95% confident to lie between $lb_c$ and $ub_c$. Because PI contains not only the variability in estimating the mean and but also randomness of individual observations, it is wider than CI.

### *B: Bootstrapping method to estimate prediction interval*

Because the interval given by conformal prediction is a kind of prediction interval, it calculates the error for each individual observation and reflects the uncertainty for both the model itself and the randomness of observations, including the impact of residuals. On the other hand, the interval obtained by bootstrapping can be interpreted as a confidence interval when it is based on the distribution of resampled model parameters or the mean of prediction values, reflecting the uncertainty of model parameters without accounting for residuals. The process of calculating confidence interval is as follows:

Assume there is a spatial prediction model $f(x)$, and a dataset containing $N$ samples. The confidence interval can be computed using bootstrapping method as following steps:

1. Randomly sample with replacement from the dataset to generate a training set $I_{train}$ of size $M$ and a test set $I_{test}$ of size $N$ - $M$.
2. Repeat the resampling process for $K$ times to create $K$ bootstrap datasets $\{I_{train,i}, I_{test,i}\}_{i=1}^{K}$.
3. Fit spatial prediction model $f(x)$ on each training set and predict the target values on corresponding test set.
4. In this way, all $N$ points can get $K$ predicted values $\{\hat{y}_{i,j}\}_{i \in I_{test}, j \in 1 \dots K}$. For each point, we can get a 90% confidence interval by taking 5% and 95% percentile $\{[q_{0.5,i}, q_{0.95,i}]\}_{i \in I}$ of its $K$ predicted values.

If we want to get the prediction interval from Bootstrapping method, we need to add random residuals $\epsilon$ to predicted values $\hat{y}$. The procedure of computing prediction intervals is a little different, which is listed as follows:

1. Randomly sample with replacement from the dataset to generate a training set $I_{train}$ of size $M$ and a test set $I_{test}$ of size $N$ - $M$.
2. Repeat the resampling process for $K$ times to create $K$ bootstrap datasets $\{I_{train,i}, I_{test,i}\}_{i=1}^{K}$.
3. Fit spatial prediction model $f(x)$ on one training set and predict the target values on corresponding test set. Also, the residuals $\{\epsilon_i\}_{i \in I_{train}}$ on one of this training sets will be obtained. Then, the final predicted values on the test set will be the predicted values plus a random error sampled from the residuals computed above $y_i^+ = \hat{y}_i + sample(\{\epsilon_j\}_{j \in I_{train}}), i \in I_{test}$. By adding this random error, we can transform the uncertainty quantification from just mean prediction to the true prediction.

4. For each point, we can get a 90% confidence interval by taking 5% and 95% percentile $\{[q_{0.5,i}, q_{0.95,i}]\}_{i\in I}$ of its $K$ predicted values.

### *C. Localized weighted exchangeability and coverage validity in the geospatial context*

**Assumption 1 (Localized weighted exchangeability).** Let $Z_i=(X_i,y_i)$ for i=1,...,n+1, and fix a measurable, nonnegative weight function supported on a neighborhood of the test point $X_{n+1}$ with:

$$w(X_i)=K\left(\frac{|l_i-l_{n+1}|}{b}\right)$$

where $K$ is a kernel and $b$ a bandwidth. We assume

$$P\big((Z_1,\dots,Z_{n+1})\in A\big)=E\left[\frac{\prod_{i=1}^{n+1} w\big(X_{\pi(i)}\big)}{\sum_{i=1}^{n+1} w\big(X_{\pi(i)}\big)}1\{(Z_1,\dots,Z_{n+1})\in A\}\right]$$

for all measurable A and permutations π. This function concentrates the mass on points $X_i$ near the test location and down-weights distant observations, thus adapting to local heterogeneity. This holds under covariate shift or when sampling is locally randomized around the test point.

**Theorem 2 (Local coverage guarantee).** Under the localized weighted exchangeability assumption, define nonconformity scores $\alpha_i=\alpha(X_i,y_i)$ and normalized weights:

$$\widehat{w}_i=\frac{w(X_i)}{\sum_{j=1}^{n+1} w(X_j)}$$

Let

$$t_{1-\alpha}(x)=inf\{t: \sum_{i=1}^{n} \widehat{w_i}1\{\alpha_i\le t\}\ge 1-\alpha\}$$

And prediction set/interval $C(x)=\{y: \alpha(X,y)\le t_{1-\alpha}(x)\}$. Then

$$P\big(Y_{n+1}\in C(X_{n+1})\big)\ge 1-$$

This result is identical in form to the global weighted case, despite weights here being local.